\documentclass[12pt, journal,onecolumn]{IEEEtran}

\usepackage{dsfont}
\usepackage{amsmath} % align
\usepackage{amsfonts}
\usepackage{graphicx}
\usepackage{multirow}
\usepackage{subfigure}
\usepackage{comment}
\usepackage{color} % to use '\textcolor'
\usepackage{setspace}
\usepackage{enumitem} % to use (1),(2),..
\usepackage{tikz}
\usetikzlibrary{positioning}
\usepackage{algorithm}
\usepackage{algorithmic}

\graphicspath{{./figures/}}

\newcommand{\vv}[1]{\ensuremath{{\mathbf{#1}}}} % Vector
\newcommand{\mm}[1]{\ensuremath{{\mathbf{#1}}}} % Matrix
\newcommand{\transpose}[1]{\ensuremath{{#1}^{\mathsf{T}}}} % Transpose
\newcommand{\bb}[1]{\ensuremath{{\mathbb{#1}}}} % for Real set
\newcommand{\E}{\ensuremath{{\mathbb{E}}}} % for Real set
\doublespace

\title{ Estimation for Compositional Data using Measurements from Nonlinear Systems using Artificial Neural Networks \vspace*{-8pt}}
\date{}

\author{ Se Un Park  
\thanks{Se Un Park is with Schlumberger, Houston, TX 77077, USA. (e-mail: seunpark@umich.edu, spark4@slb.com). }
}

\begin{document}

\maketitle

\begin{abstract}
Our objective is to estimate the unknown compositional input from its output response through an unknown system after estimating the inverse of the original system with a training set.
The proposed methods using artificial neural networks (ANNs) can compete with the optimal bounds for linear systems, where convex optimization theory applies, and demonstrate promising results for nonlinear system inversions. We performed extensive experiments by designing numerous different types of nonlinear systems.
\end{abstract}

\section{Introduction}
Compositional data is used in many fields because the data in population ratios or fractions is easy to interpret. However, when the compositional data cannot be produced by simple scaling or normalization with the whole population size from the raw data or measurements, the process to produce such compositional outputs may not be straightforward. Here, we consider noisy outputs as our observations from an unknown linear or nonlinear system with the corresponding compositional variable inputs of interest. The pairs of input and outputs will be used as a training set for artificial neural networks (ANN) modeling to estimate the inverse of the unknown system. This trained inverse system can predict the unknown compositional input, given the output measurement coming from the original system with the input. 
As our approach is based on ANNs, we do not directly estimate the forward observation model, as in the classical inversion theory, but the inverse of the original system. The measurements, the outputs from the original system with the compositional inputs, are then the input of our estimated inverse system, which will predict the original compositional inputs. 
We do not apply post-processings or ad-hoc approaches such as truncation of the estimate followed by scaling so that the final answer is a non-negative vector that sums up to one. Rather, we directly apply non-negativity and scaling layers in the proposed ANNs. 

We considered both linear observation models and several types of nonlinear models. For the linear cases, where we can theoretically analyze the optimal performance bounds, we 
demostrated with our experiments that the performance of ANNs for the inversion of the linear model outputs can compete with the optimal bounds.
For the nonlinear systems, where convex optimization methods are not well suited for these general cases, we could still present promising results compared to the error levels in the linear models and leave the comparitive analysis with other feasible optimization methods for our future work.

\section{Observation Models}
\label{sec:models}

We first define a compositional vector and then present a general observation model. Then, we will formulate more specific observation models. An example of a compositional data or vector includes population ratos, concentration of chemicals in the air, numerous survey statistics in percentage. 

We define the compositional vector $\vv{m}$ to be constrained such that its components are nonnegative and sum to unity. 
These constraints define a simplex set such that any compositional vector \vv{m} is in the simplex set. 
An $M$-dimensional simplex, or simply $M$-simplex, is defined by 
\begin{equation}
\label{eq:def_simplex}
S^M = \{ (x_1, \ldots , x_M) \in \bb{R}^M \, : \,  \sum_{i=1}^M x_i = 1, x_i \geq 0 \mbox{ for } \forall i \} .
\end{equation}

Let $m_i$ be the $i$th component of a compositional column vector $\vv{m}$, then it can be denoted by $\vv{m} = \transpose{[m_1, m_2, \dots, m_M]}$ where  $\mathsf{T}$ is a transpose operator. Further decomposing it leads to $\vv{m} = \sum_{i=1}^M m_i \vv{e}_i$in terms of its components with basis vectors $\vv{e}_i$, which is $i$th column of $M \times M$ identity matrix $\mathbb{I}_M$.

We now assume the following system $h$, i.e., a forward, observation model that generates our observation \vv{s} from the $M$ dimensional compositional input \vv{m} such that $\vv{m} \in S^M$.
\begin{equation}
\vv{s} = h (\vv{m}) + \vv{n},
\end{equation}
where $\vv{s} \in \mathbb{R}^L$, $h$ is a function from $S^M$ to $\mathbb{R}^L$, and \vv{n} is additive noise\footnote{For multiplicative noise, taking log transformation of the observation leads to the same formula.}.

In the rest of this chapter, we define specific forms of a nonlinear system $h$ with more restrictions as we proceed, finally leading to a linear model.

\subsection{General Systems}
The system response from an input \vv{m}, without noise, is 
\begin{align}
h(\vv{m}) &= h\left(\sum_{i=1}^M m_i \vv{e}_i \right) .
\end{align}
The input \vv{m} is decomposed by using trivial bases $\vv{e}_i$s. 
If the system behaves nonlinearly or non-parametric ways without closed forms, 
 then for the characterization of the system and the inversion for the input with the given the output, mapping or non-parametric esimations such as based on nearest neighbors of pairs of input and output could be working solutions. Training of ANNs is also possible as a candidate mapping solution.

For example, $h(\vv{m}) =  \mm{A}^{p(\vv{m})} \vv{m} \times \exp ( - K \| \mm{B} \vv{m} \|_2 ) $ where $p(\vv{m}) = ceil( \mm{C} \vv{m} ) $, $ceil(\cdot)$ is a ceiling operator that maps to integer domain, dimensional compatible matrices \mm{A,B,C} and, a scalar constant $K$.

\subsection{Systems with additivity}

\subsubsection{A System with partial additivity}

If the system holds partial additivity for several sets of groups, $G_k$s, each of which is a set of component indices for the input vector \vv{m}, then
\begin{align}
h\left(\sum_{i=1}^M m_i \vv{e}_i \right) = \sum_{k} h'_{G_k} \left(\sum_{i \in G_k} m_i \vv{e}_i \right) =  \sum_{k} h'_{G_k} \left( \{ m_i \}_{i \in G_k} \right),
\end{align}
where $h'_{G_k}$ is a function of the same dimension as $h$ but specific to the group $G_k$ and $ \{ m_i \}_{i \in G_k}$ is a tuple of the components of \vv{m} in the indices in $G_k$. Note that $G_k$s do not have to be exhaustive such that the intersection of $G_k$ and $G_j$ for $i \neq j$ may not be empty. 

A special case of this system can be the multiplicative system with the constant vectors $\vv{h}_k$ corresponding to $k$th functions of $\vv{m}$, $g_k(\vv{m})$. 
This can be seen as a linear system with respect to $g_k(\vv{m})$s.
\begin{align}
\label{eq:lin_g}
h\left(\sum_{i=1}^M m_i \vv{e}_i \right) = \sum_{k} \vv{h}_{k} g_k(\vv{m}),
\end{align}
where $\vv{h}_{k} \in \bb{R}^L $ is a constant vector and independent of \vv{m} and $g_k$ is a nonlinear scalar function of \vv{m}.
Note that $g$ can be either invertible or non-invertible. For the special case of the latter, where $g$ is a thresholding operator, we can minimize the inevitable estimation bias by configuring the optimal (inversion) mapping rule from output to input. Refer to Appendix  \ref{appen:noninvertible_rule}.

This special case model can be practical because a general system on a simplex $S^M \subset [0,1]^M$ can be well-approximated if $h$ is differentiable with Taylor expansion. Even non-differentiable systems can be approximated to differentiable ones and can be decomposed. For a point $\vv{a} \in S^M$, a general system response $h(\vv{m}) \approx \sum_{k} \vv{h}_{k} g_k(\vv{m}) $ with $\vv{h}_{k} = D^\alpha h(\vv{a}) , g_k(\vv{m}) = (\vv{m}-\vv{a})^\alpha  / \alpha !$, for order $\alpha$, but note that the notation $g_k$ is `loosely' defined in relating the order $\alpha$. For example, for $\vv{m} \in S^2, \vv{a}=\vv{0}, \alpha=2$, $g$ in the $k$th term can be either $m_1^2/2, m_2^2/2$, or $ m_1 m_2$. For more precisely defined terms, refer to Appendix \ref{appen:TaylorExpansion}.

Example of this model can be as the followings: 

$h(\vv{m}) = \sum_{k} \vv{h}_{k} g_k(\vv{m})$ with $g_k(\vv{m}) = m_k m_{k+1}$ for $k \in [1,M-1]$ and $g_M = m_M m_1$.

$h(\vv{m}) = \mm{A} \vv{m} \times \exp ( - K \| \mm{B} \vv{m} - \mu\|_2^2/2 ) $

$h(\vv{m}) = \mm{H} \vv{g}(\vv{m})$ with $\vv{g}(\vv{m}) = \transpose{[m_1 , 0.4 m_2, 0.2 m_1^2 , m_3^2 , 0.7 m_1 m_2 ]}$

\subsubsection{An additive system with component-wise responses $ \vv{h}_i$}

If additivity holds for the system and the component-wise system response depends on the composition, then we model this system as the following. 
 
\begin{align}
h\left(\sum_{i=1}^M m_i \vv{e}_i \right) = \sum_{i=1}^M h \left( m_i \vv{e}_i \right) = \sum_{i=1}^M \vv{h}_i(m_i), 
\end{align}
where $\vv{h}_i(m_i) \in \mathbb{R}^L$ is a function of a scalar $m_i$. 
For the $i$th component, the system response $ \vv{h}_i$ depends on the composition of $m_i$, such as shape change in the response. 
For example, 
$$\vv{h}_i(m_i) = \left( ( a_1 -a_0 ) m_i + a_0 \right) \exp\left( -K \left( \vv{x}  - \left( (b_1 - b_0) m_i + b_0 \right) \right)^2 \right) $$ for a fixed index vector in observation $ \vv{x} =\transpose{[1,..., L]}$. The peak location of this function is translated from $b_0$ to $b_1$ and the magnitude of the peak is scaled from $a_0$ to $a_1$, as $m_i$ changes from 0 to 1.

\subsubsection{An additive system with fixed-shape component-wise responses $ \vv{h}_i$ and nonlinear scaling factors}

If additivity holds for the system and the component-wise system response is a scaled version of a fixed shape characterized by the component, then we model this system as the following.
\begin{align}
\label{eq:system04}
h\left(\sum_{i=1}^M m_i \vv{e}_i \right) = \sum_{i=1}^M h \left( m_i \vv{e}_i \right) =  \sum_{i=1}^M g_i(m_i) h(\vv{e}_i) =  \sum_{i=1}^M g_i(m_i) \vv{h}_i, 
\end{align}
where $g_i(m_i)$ is an arbitrary scalar function on the specific component of index $i$ and $\vv{h}_i = h(\vv{e}_i) \in \mathbb{R}^L$. For example, $g$ can be quadratic or piecewise continuous: $g_1(t) = t^2, g_2(t) = t^{0.3} $ where $t \in [0.2 , 0.3]$ and zero elsewhere.

\subsubsection{A Linear System}

When linearity holds for the system response, then $g$ from \eqref{eq:system04} can be treated as an identity operator, i.e., $g(t)=t$ and
\begin{align}
\label{eq:linearModel_noiseless}
h\left(\sum_{i=1}^M m_i \vv{e}_i \right) = \sum_{i=1}^M m_i \vv{h}_i = \mm{H} \vv{m},
\end{align}
where \mm{H} is a linear system matrix comprising $\vv{h}_i$ as its $i$th column.
The analysis and inversion under this linearity assumption were performed in our previous work \cite{Park2019spe}.

\subsection{Systems with missing or noise compositional vector as obfuscating unknowns }
Here, we do not assume a complete knowledge of the dimension $M$ of the unknown compositional vector but we are ignorant of a partial vector in some dimensions or interested in the compositional vector except this partial vector. In other words, we consider that the whole compositional vector $\vv{m}^0$ comprises two components 
${\vv{m}}, {\vv{m}^1}$ 
and the measurement forward model is 
\begin{align}
\label{eq:sys_obf}
\vv{s} = h (\vv{m}^0) + \vv{n} = h \left(\left[
\begin{array}{l}
\vv{m} \\
\vv{m}^1 
\end{array}
\right] \right) + \vv{n} .
\end{align}
We assume that we do not have knowledge of the existence of the obfucsticating unknown vector or compositional noise vector $ \vv{m}^1$ and equivalently we are interested in obtaining only \vv{m}. 
The training set consists of pairs $ (\vv{m}_i, \vv{s}_i)$ without $\vv{m}^1$. 
In practice, such a compositional noise vector $\vv{m}^1$ can be from environmental effects, which are difficult to measure but still affects -- even controlled -- experiments.

Note that this model includes a trivial but practical case where a constant bias is added to $h(\vv{m})$ and our observation, eg., spectral offsets from environments such as contribution of environmental elements in X-ray based spectroscopy.

\section{Baseline Performance Analysis for Inversion }

Considering the models introduced in the last chapter, we will provide analyses based on the loss functions, metrics, and obfuscating variables in this chapter. 
Because the inversion performance of nonlinear systems with the simplex constraint is difficult to analyze compared to the linear inversion without the constraint, we provide theoretical analysis or bounds for the linear case as surrogate ones.

\subsection{Loss functions and performance metrics}

\subsubsection{Loss function with the compositional target}

Ideally, we want to directly minimize some distance, as an estimatior error, between the estimate and the true composition vector. 
In other words, 
the loss function in an ideal form can be used to minimize a distance $d(\cdot,\cdot)$ between the true vector $ \vv{m}_{true}$ as the target and the estimated vector $\hat{\vv{m}} =  f(\vv{s})$ obtained from an estimator $f$ on the corresponding measurement \vv{s}, as seen below.
\begin{align}
\label{eq:ideal_loss}
L_{ideal} = d (  \vv{m}_{true} , \hat{\vv{m}} ), 
\end{align}
where both $\vv{m}_{true}$ and $\hat{\vv{m}}$ satisfy the simplex constraints.

A trained system after minimizing \eqref{eq:ideal_loss} using a set of samples $\{\vv{m}_i,\vv{s}_i \}_i$ can produce a compositional estimate on a new measurement but this estimation is performed through mapping of the measurement as an input to the system, not by typical inversion.
In this work, we will perform the optimization of the mapping function $f$ by minimizing the above distance using ANN on the training set under a given model order or hyper-parameters. The trained model retains estimated parameters such as weights and biases.

Considering possible convex optimization approaches, we note that it is difficult to formulate and efficiently solve a convex loss function with an explicitly form of $f$ because of the simplex conditions. 
For example, the typical projection onto a simplex is not a convex function. 
The simplex constaint is linear but applying the boundary conditions is not always trivial, especially in high dimensional space \cite{Park2019spe}. To the best of our knowledge, efficient convex optimization algorithms guaranteeing global optimal solutions are difficult to find. In contrast, ANNs are generally non-convex with nonlinear activation functions but its training phase, if performed well with sufficient data, empirically guarantees good performances with a large size of training samples.

\subsubsection{Loss function with the measurement}

In practice or testing of the inversion of a measurement \vv{s} by using the trained system, we cannot directly minimize the distance of the estimate from the true compositional input because the input is not known but will be estimated. Therefore, many inversion methods do not use the ideal loss function of \eqref{eq:ideal_loss} with the unknown \vv{m} but adopt loss functions of the measurements and the estimated projections on the observation domain, called projection errors.
For practical optimization using measurements only, we will use the following loss function 
\begin{align}
\label{eq:loss_surrogate}
L = \| \vv{s} -  T(\vv{m}) \| ,
\end{align}
with $ \vv{m} \in S_M$ and $\ell_2$ distance $\| \cdot \|$.

The simplest case of this type of optimization is for the linear system and the unconstrained domain for \vv{m}, i.e., $T(\vv{m}) = \mm{A} \vv{m} $ for $ \vv{m} \in \bb{R}^M$. Standard, classical linear regression methods can be used for this unconstrained optimization in minimizing the distance between the linear observation and the projection of the estimate. 

We note a special case where training samples are used for the estimation of linear system with the simplex constraint \cite{Park2019spe}. This work cannot cover nonlinear systems but shows how the direct inversion is effectively done after training the linear system having compositional inputs as unknowns.

For this simplest case with linear systems $T$, 
in the view of approaches using ANNs,
the minimum structure is a shallow network where only one matrix of weights without bias is used. This weight matrix is the same as the pseudo-inverse of the linear system matrix \mm{A}, denoted by $\mm{A}^\dagger$. However, we empirically confirmed that the ANN with this minimum order converges slowly but higher ordered models converge fast while guranteeing the performance. Such higher orders seem redundant at first but we experimentally observe that they converge and perform better and consistently throughout our different experiments. In other words, the minimum possible structure in ANNs may not be the practically optimal. We adopted this principle in our work.

\subsubsection{Performance metrics}
\label{sssec:metrics}

For fair comparisons of different methods, we use the following metrics of $e$ (average of $l_2$ distances of errors) and  $aad$ (average of absolute deviations or errors) in percent (\%).
\begin{align}
\label{eq:l2error}
e &= \frac{1}{N} \sum_i^N \| \vv{m}_{i,true} - \hat{\vv{m}}_i \|_2 \times 100 , \\
\label{eq:l1error}
add &= \frac{1}{N} \sum_i^N | \vv{m}_{i,true} - \hat{\vv{m}}_i | \times 100 ,
\end{align}
where $N$ is the sample size and $|\vv{x}|$ is a vector of component-wise absolute value, i.e., $[|x_1|, ..., |x_M|]$

\subsection{Benchmark performance in linear systems}

\subsubsection{Inversion with the knowledge of the dimension of unknowns}
Here, we assume a linear system $T=\mm{H}$ to produce closed form metric as a (surrogate) benchmark performance. Also, we assume a complete knowledge of the dimension $M$ of the unknown compositional vector. 
We assume that \mm{H} is full-rank and overdetermined, $M < L$, so $\transpose{(\mm{H}} \mm{H}) ^{-1} $ is well defined.

Let  $\mm{H} = \mm{U} \mm{S} \transpose{\mm{V}}$ by singular vector decomposition and $diag(\mm{S}) = \transpose{[ s_1, ..., s_M]}$, where $diag()$ is an operator that vectorize a matrix by extracting diagonal entries.
let $ \mm{H}^\dagger := (\transpose{\mm{H}} \mm{H}) ^{-1} \transpose{\mm{H}} $ be the pseudo inverse of $\mm{H} $.

The expected error in $\ell_2$ norm, $d_{oracle,uc}$, on unconstrained domain for $\vv{m}$ is calculated as follows.
\begin{align}
\label{eq:oracle_est_error2}
d_{oracle,uc}^2 &= \E \| \vv{x} - \mm{H}^\dagger \vv{y} \|^2 
= \sigma^2  \sum_k^M s_k^{-2}, 
\end{align}
where $tr()$ is the trace operator. Therefore, the equation \eqref{eq:loss_surrogate} becomes 
\begin{align}
\label{eq:oracle_est_error}
d_{oracle,uc} = \sigma  \sqrt{\sum_k^M s_k^{-2} }, 
\end{align}

\subsubsection{Inversion with missing or noise compositional vector as obfuscating unknowns }

If we know that there can be obfusating variables, then the standard simplex constraint for the estimated portion $\vv{m}$ should be relaxed; we will have sum-to-less-than-or-equal-to-1 constraint instead of sum-to-one.
Without knowing the dimension of missing or obfuscating variables, or simply our ignoring such variables, we can re-define the estimation error for a composition vector $\hat{\vv{m}} \in S^M$ with the partial true vector \vv{m} of interest but without the noise vector $ \vv{m}^1$ in \eqref{eq:sys_obf}, by normalizing \vv{m} so that it satisfies the simplex constraint.
\begin{equation}
\label{eq:ideal_loss_obs}
L_{ideal}^2 = \| \frac{\vv{m}}{\|\vv{m}\|_1} -  \hat{\vv{m}} \|_2^2 
\end{equation}

We provide an analysis of impact of an obfuscating vector on inversion for linear systems. The observation model equation can be rewritten as 
\begin{align}
\vv{s} = h (\vv{m}^0) + \vv{n} =  \|\vv{m}\|_1 \mm{H} \left( \frac{\vv{m}}{ \|\vv{m}\|_1} \right) + \mm{H}^1 \vv{m}^1 + \vv{n} =  \vv{H}'\vv{m}' + \vv{n}' ,  
\end{align}
where $ \mm{H}' = c \mm{H} , c=\|\vv{m}\|_1 \in (0,1], \vv{m}' = \vv{m} / \|\vv{m}\|_1 \in S^M, \vv{n}' = \mm{H}^1 \vv{m}^1 + \vv{n}$. Therefore, {\it in practice without the knowledge of even existence of an obfuscating vector of missing variables, we seek a solution in a simplex where the linear system matrix is scaled with also an unknown factor $\|\vv{m}\|_1$ from a measurment mixed with perturbed noise $\vv{n}'$}.  The effective noise $\vv{n}'$ is generally centered at a non-zero vector and even correlated, even if $\vv{n}$ is zero-mean and uncorrelated because of the unknown system $\mm{H}^1$ and the obfuscating vector $\vv{m}^1$. The obfuscating vector can be treated as either a fixed unknown or a stochastic quantity which leads to correlated effective noise $\vv{n}'$.

The loss $L$ is defined as the following. 
\begin{equation}
L^2 = \| \vv{s} -  \mm{H}\hat{\vv{m}} \|_2^2 
\end{equation}

Without knowing $\mm{H}^1$, to obtain $\hat{\vv{m}}$, a `myopic' estimator uses only \mm{H}, which is either given or estimated. A simple myopic estimator is 
$ \hat{\vv{m}} = P ( \mm{H}^\dagger \vv{s} ) $, 
where $P(\vv{x}) = P_s( P_t(\vv{x}))$ projects any nonzero vector $\vv{x} \in \mathbb{R}^M$ to $S^M$,  $P_t(\vv{x}) = [ .. \max(0,x_i) ..] $ is a thresholding opereator, $P_s(\vv{x}) = \vv{x} / \|  \vv{x} \|_1$ is a scaling operator.

The expected squared loss with an unconstrained pseudo-inverse of $\mm{H}$ without projection $P$ is
\begin{align}
\E L^2 &= \E \| \vv{s} -  \mm{H}\hat{\vv{m}} \|_2^2 = \E \| \mm{P}_\perp \mm{H}^1 \vv{m}^1 + \mm{P}_\perp \vv{n} \|^2 = \| \mm{P}_\perp \mm{H}^1 \vv{m}^1 \|^2 + \E \| \mm{P}_\perp \vv{n} \|^2 \\
&= \| \mm{P}_\perp \mm{H}^1 \vv{m}^1 \|^2 +  \sigma^2 tr( \mm{P}_\perp)
= \|  \transpose{\mm{U}_1} \mm{H}^1 \vv{m}^1 \|^2 + (M' - M) \sigma^2 \\
& \leq  \|\vv{m}^1\|_1 \lambda( \transpose{\mm{U}_1} \mm{H}^1 ) + (M' - M) \sigma^2 , 
\end{align}
where $\mm{P} = \mm{A}\mm{A}^\dagger$ is a projection matrix of \mm{A}, 
$\mm{P}_\perp = \mm{I} - \mm{P} = \mm{I} - \mm{U} \transpose{\mm{U}} = \mm{U}_1 \transpose{\mm{U}_1}$ is a orthogonal projection matrix of \mm{A}, $\mm{A} = \mm{U} \mm{S} \transpose{\mm{V}}$  by SVD,  $\mm{U}_1$ have orthogonal basis vectors with which \mm{U} span $\mathbb{R}^{M'}$ with $M'$ being the sum of the dimensions of $\vv{m}$ and $\vv{m}^1$ ($\vv{m}^0 \in \mathbb{R}^{M'}$), $\vv{n}$ follows Gaussian distribution with mean zero and covariance matrix $\sigma^2 \mm{I}$, $\lambda(\mm{A})$ is the largest eigenvalue of \mm{A}, and $tr(\cdot)$ is a trace operator.

The squared estimation error is 
\begin{align}
\label{eq:obfuscated_est_error2}
 \E \| \vv{x} - \hat{\vv{x}}\|^2 & = \E \| \mm{H}^\dagger \vv{n}' \|^2 =  \| \mm{H}^\dagger \mm{H}^1 \vv{m}^1 \|^2 + \E\| \mm{H}^\dagger \vv{n} \|^2 = \| \mm{H}^\dagger \mm{H}^1 \vv{m}^1 \|^2 + \sigma^2  \sum_k^M s_k^{-2} \\
& \leq  \|\vv{m}^1 \|_1 \lambda^2(\mm{H}^\dagger \mm{H}^1) + \sigma^2  \sum_k^M s_k^{-2} . 
\end{align}
This error has an additional obfuscating factor $\|\vv{m}^1 \|_1 \lambda^2(\mm{H}^\dagger \mm{H}^1)$ compared to \eqref{eq:oracle_est_error2}. We note that this error converges to \eqref{eq:oracle_est_error2} when the obfuscating variables become negligible ($\vv{m}^1 \rightarrow \vv{0} $) or the system for the variables has a negligible effect ($\mm{H}^1 \approx \mm{0}$).

\section{Experiments}
We perform experiments based on the examples following the described models in Section \ref{sec:models}. We start from the simple models %st model -- linear models -- 
to more complex and nonlinear models. 

\subsection{Design and implementations}

We implemented the designed simulations using Python 3.5 and  extensively experimented several objective functions, strucutures, tuning strategies, and different nonlinear and non-negative activation functions in ANNs. 

First,
to efficiently train ANNs and to better generalize, we include some redundancy in the structure. Indeed, minimal structures may not guarantee good convergence rate or sometimes fail to converge due to sensitivity, e.g., linear systems and modeling of it using only weights linking input and output directly. 
Further redundancy to avoid overfitting such as dropout layers was tried but not used in our experiments because they did not improve the estimation or has little effect. Batchnorm layers are inserted between layers for efficient training.

To obtain compositional vectors as outputs of our estimators, we added a simplex projection to the last layer in our ANNs, which is nonconvex. Here, we apply only rescaling of the vector, by dividing it with the sum of the vector components obtained from the previous layer, because the chosen activation function of the layer already guarantees non-negativity. 
We note that optimization of ANNs is a generally non-convex procedue but with rich empirical guidelines to avoid local minima and achieve satisfactory performance.

As an objective function to minimize, 
we use a mean squared ($\ell_2$) distance between the ANN output $:=ANN(\vv{m})$ and \vv{s} in the loss function to optimize the ANNs, after trying different distances such as mean abosolute distance (using $\ell_1$ distance), mean absolute percentage distance, categorial crossentropy, soft-max types, etc. We empirically confirmed that using the $\ell_2$ distance achieves the performance in terms of lowest estimation bias and fast convergence rate. 

Among many optimizers or packages,
we adopted Adam optimizer for ANN training \cite{Kingma2014} after experimenting other optimizers such as SGD, RMSProp, Adagrad, Nadam in Keras package \cite{Keras2015}. Also, we have tried many tuning strategies and the tunned parameters are mostly default values: $\beta_1=0.9, \beta_2=0.999$, decay rate is 0.01. The learning rates and batch sizes depend on the experiments and range from $10^{-6}$ to $10^{-3}$, from 64 to $N_{training}$, respectively. In training stages, we checked the validation errors so that the overfitted parameters are not used in testing.

We evaluated the performance mainly using the compositional samples drawn according to the uniform distribution in a simplex, because this distribution is the most scattered distribution having the highest entropy in information theory under the volume measure. However, we added several tests having compositional samples drawn according to a mixture of concentrated distributions and uniform distribution.

\subsection{Simple linear systems}
We perform the experiments on the linear systems of the low dimensional spaces of observations and unknowns. We set $L = 5, M = 3, N = 10000$ (the number of training samples), 
$N_{test} = 10000$
 (the number of testing samples). Thus, even if we do not know the system function, the multiplicative system matrix in this case, we know its dimension and the matrix will be estimated using the training data.

We simulated the linear system matrix \mm{H} so that each of its entries was generated according to standard Gaussian distribution.
The training and test set of compositional vectors are generated uniformly on simplex $S^L$ \cite{Onn2009}. Let $\mm{X}_{train}$ and $ \mm{X}_{test}$ be the matrix comprising of the true label (compositional) vectors in the training and test set, respectively.

The realistic linear model can be described with an additive noise as the following
\begin{align}
\label{eq:linearModel}
\vv{s} = \mm{H} \vv{m}  + \vv{n} ,
\end{align}
where \vv{n} is a noise vector.  The additive noise vector in \eqref{eq:linearModel} is generated such that each entry of the vector follows mean zero Gaussian distribution with standard deviation $\sigma = 0.005$. 
The system responses in the training and test sets using the compositional input $\mm{X}_{train}$ and $ \mm{X}_{test}$ are collected into the matrix $\mm{Y}_{train}$ and $ \mm{Y}_{test}$, respectively. The MLE (maximum likelihood estimator) of the system matrix is obtained as the following \cite{Park2019spe}:
\begin{align}
\label{eq:MLE_H_linear}
\hat{\mm{H}}=\mm{Y}_{train}\transpose{\mm{X}_{train}}(\mm{X}_{train}\transpose{\mm{X}_{train}})^{-1} .
\end{align}
Using such an estimated linear system matrix, we perform inversion to estimate the unknown compositional vector from its system response. 

For the experiment with ANNs, we try two cases: ANN with one layer vs. ANN with multiple layers.
We measure the estimation performance by evaluating the difference of matrix of the test set $ \mm{X}_{test}$ and the matrix of the estimated compositional vectors obtained from $\mm{Y}_{test}$. The error metric is precisely formulated by equation \eqref{eq:l2error} in Section \ref{sssec:metrics}. 

We note that the shallowest ANN will have nonunique optimal solutions depending on initialization or randomization. This is described in Appendix.~\ref{app:nonuniqueness_shallow} and we do not experiment on this shallow structure.

\subsubsection{ANN with 1 layer}

We present several trivial ANN learning cases to demonstrate that our intuitions match the desired behaviors of the learned models. We omit reporting error values of these trivial cases.
We first train this shallow ANN to learn the mapping from compositional domain in $S^M$ to the output domain. The learned ANN should have the weight matrix related to the original linear system matrix. We below provide the discussion of this considering an optional bias term in ANNs and both forward and inversion models.

\begin{itemize}
\item{ Estimation of linear system matrix \mm{H} without a bias term: }
We model $ANN(\vv{m}) \approx \vv{s}$. The input $\vv{m} \in S^M$ is multiplied by the first ANN weight matrix $\mm{W}_1$ and the distance between this vector and the desired system output \vv{s} is minimized. 
We experimentally observed that the trained mapping result was good, i.e., $ANN(\vv{m}) \approx \vv{s}$  and the weight matrix $\mm{W}_1 \approx \hat{\mm{H}}$ as expected. 

\item{ Estimation of linear system matrix \mm{H} with a bias term: }
The input $\vv{m} \in S^M$ is multiplied by the first ANN weight matrix and added with a bias term. 
We empirically obtained the same good results as above but, the weight matrix $\mm{W}_1$ differs the system matrix \mm{H} and the MLE $\hat{\mm{H}}$ because of the bias term in the ANN. Theoretically, if the distribution of the training samples cover all the possible domain space and $N$ goes to infinity, the bias terms will converge to zero and $\mm{W}_1 \rightarrow \hat{\mm{H}}$. 

\end{itemize}
The above cases consider learning the forward model whereas the below cases consider learning the inversion so that the ANN can produce the compositional vector \vv{m} from a measurement \vv{s}.

\begin{itemize}

\item{(Inversion) Estimation of pseudo-compositional vector without a bias term: }
Similar to matrix inversion, we used a linear activation function after mulitplying a weight matrix. The trained ANN performs good inversion and the result is comparable to using the inverse matrix of the estimated \mm{H}, i.e. $\hat{\mm{H}}^{-1}$. Thresholding and scaling operations are required to project the ANN output onto the simplex domain. 

\item{(Inversion) Estimation of pseudo-compositional vector with a bias term: }
Similar to the above case, we used a linear activation function after mulitplying a weight matrix but adding a bias. The trained ANN performs good inversion and the result is comparable to using the inverse matrix of the estimated \mm{H} but with a constant term due to the introduced bias term in the model. Thresholding and scaling operations are required to project the ANN output onto the simplex domain. 

\item{(Inversion) Estimation of compositional vector without a bias term: }
We performed a similar experiment as above but added a mapping layer so that the ANN ouput is in a simplex.  Then we do not need to apply thresholding and scaling operations to project the ANN output onto the simplex domain as done above. Throughout experiments\footnote{The softmax activation was not good in training for this shallow layered ANN in our experiments even with batch normalization after weight multiplication.}
 we could observed that this ANN shows good performance without a need for post-processing of mapping onto a simplex.

\end{itemize}

From the observation of the above last case demonstrating good inversion with a projection layer, we can extend the model further by adding another layer before the projection.

\subsubsection{ANN with multiple layers}
\label{sssection:linear}

To investigate extendibility of ANNs with muliple, possibly deep, layers, we designed two-layered ANN with the projection layer as the last layer. The first and second layers each have $4 \times M$ nodes, each followed by batch normalization and applying a sigmoid activation, and the last layer has $M$ nodes with ReLu activation \cite{Nair2010} followed by the scaling operation as the projection layer because non-negativity is guaranteed by the previous activation function. 

Note that the generated system matrix can have negative numbers, as was in our realization that was used throughout in our applicable experiments with its condition number 3.23 (the ratio of the largest singular value to the smallest). %
condition number

The errors of \eqref{eq:l2error} are 
\begin{align*}
e_{oracle} &= 0.56962239814362881 ,\\
e_{benchmark} &= 0.56958924114641452 ,\\
e_{ann} &= 0.5716387941750467 ,
\end{align*}
where the $oracle$ case uses the true system matrix for inversion so the estimator is $P(\mm{H}^\dagger \vv{s}) \in S^M$, $benchmark$ case uses the MLE of the system matrix for inversion so the estimator is $P(\hat{\mm{H}}^\dagger \vv{s})$, and $ann$ case indicates results from the trained ANN.
The three error values are comparable. The error from the ANN approach is slightly larger than the rest.

The difference between $\mm{H}$ and $\hat{\mm{H}}$ is 
$$ \| \mm{H} -\hat{\mm{H}} \|_F / \| \mm{H}  \|_F = 0.00016788000456487503  , $$
where $\|\cdot\|_F$ indicates a Frobenius norm. This small number implies the MLE for the system matrix is accurate enough and the benchmark performance with MLE should be similar to the oracle case, as shown above.

We note that the theoretical bound for unconstrained estimator \eqref{eq:oracle_est_error2} is 
$$ d_{oracle,uc} \times 100 = 0.68009831763330508 . $$ 
This is significant larger than the error level of 0.57 seen in $e_{oracle},e_{benchmark},e_{ann}$ we obtain from several estimators, primarily because the simplex constraint applied with a projectiion operator or scaling seems to limit the variable ranges unlike the unconstrained estimator\footnote{We performed our experiments multiple times with different randomized-realization of the system matrix so this trend of observation is valid.}.

\subsection{Simple nonlinear systems}

We perform the experiments on several different nonlinear systems of the low dimensional spaces of observations and unknowns. Most of these have dimensions of $L = 5, M = 3$, unless explicitly stated, and $N = 10000$ (the number of training samples), 
$N_{test} = 10000$  (the number of testing samples).

\subsubsection{Nonlinear systems: invertible transformation on simplex variable}
\label{sssection:nonlinear_inv}

We designed a nonlinear system where the output should be uniquely invertible to the original input without noise. We designed the following particular nonlinear system:
\begin{align}
T(\vv{m}) &= \mm{H} \, g(\vv{m}), \\
\label{eq:g_nonlinear_inv}
g(\vv{m}) &= \transpose{[  m_1 ^2 , m_2^{0.5} + 0.1 , m_3 ]},
\end{align}
where \mm{H} has entries generated according to the standard Gaussian distribution.

The inverse function of $g$ is as the following:
\begin{align}
\label{eq:g_inv_nonlinear_inv}
g^{inv}(\vv{x}) &= \transpose{[  ( P_t(x_1) ) ^{0.5} , ( P_t(x_2 - 0.1) ) ^{2} , x_3 ]},
\end{align}
where $\vv{x}$ is not necessarily in a simplex and can be negative as an input argument of $g^{inv}$ due to the presence of noise, thus requiring non-negative projection $P_t$ for the square-root operation, and the third variable is by-passed as in $g$.

The averaged $\ell_2$ errors in percentage are, again for $N_{test}=10000, \sigma=0.005$,
\begin{align*}
e_{oracle} &= 0.899552 ,\\
e_{benchmark} &= 0.899289 ,\\
e_{ann} &= 0.493694 ,
\end{align*}
where the $oracle$ case uses the true system matrix for inversion so the estimator is $P(g^{inv}(\mm{H}^\dagger \vv{s})) \in S^M$, $benchmark$ case uses the MLE of the system matrix for inversion so the estimator is $P(g^{inv}(\hat{\mm{H}}^\dagger \vv{s}))$, and $ann$ case indicates results from the trained ANN but without knowledge of $g$.
It is surprising to note that ANN significantly beats other two estimators. 
We may not directly compare the results coming from two different systems of this nonlinear system and the previous linear system. However, it is clear to notice the gap of errors from ANN and the pseudo-inversion methods, compared to the plain linear model in \ref{sssection:linear} with the negligible gap in errors from different methods. 
 The only change added to the linear model is the additional nonlinear effects on $m_1, m_2$ by the function $g$. Again, the benchmark case is similar to the oracle case because of close proximity of $\hat{\mm{H}}$ to $\mm{H}$. 
It is noteworthy to observe that the performance of these two has relatively degraded due to nonlinear effects of $g$, %, which is the only change the linear system, 
while the ANN performance relatively improved even without using functional form of $g$. 

This result also implies that there must be optimal estimator better than the above `oracle' estimator, which should depend on a particular nonlinear function $g$. The cascading inversion operation after the pseudo-inverse with the system matrix may better be combined but the search of the better estimator, although interesting, is not in the scope of this work and we leave it as future work.

\subsubsection{Nonlinear systems: noninvertible transformation on simplex variable}
\label{sssection:nonlinear_noninv}

Unlike the previous experiment above, 
we consider partially noninvertible and nonlinear transformation on simplex variables. Because of partial noninvertibility, the estimation has an unavoidable bias regarding the noninvertible space. 

\begin{figure} [h]
 \centering %  trim={<left> <lower> <right> <upper>}
 \includegraphics[width=4in,trim={0cm 0cm 0 0cm},clip]{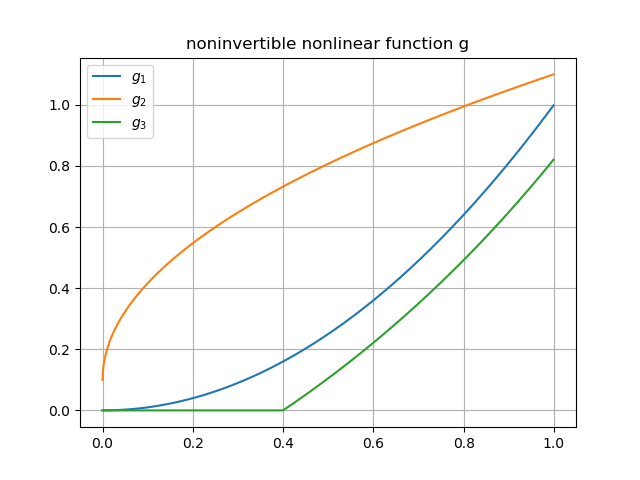}
 \caption{\label{fig:g_nonlinear_noninv} An example of $g$ of \eqref{eq:g_nonlinear_inv} for $m_1,m_2$ and \eqref{eq:g_3} for $m_3$. This tranformed vector from a simplex vector is multiplied by matrix \mm{H} and noise is added to it to synthesize an observation vector. 
 }
\end{figure}

In our experiment, we apply $g(\cdot)$ and $g^{inv}(\cdot)$ of equations \ref{eq:g_nonlinear_inv} and  \ref{eq:g_inv_nonlinear_inv}, respectively, which perform transformations on the first two dimensions of \vv{m}. We added a noninvertible transformation with a thresholding operator on $m_3$ as below. 
\begin{align}
\label{eq:g_3}
g_3(x) &= \exp( P_t(x -T) ) -1 , \\ 
g_3^{inv}(x) &= 
\begin{cases}
    x',  & \text{if } x' = \log P_{t,\epsilon}(x+1)'\geq T\\
    T/2, & \text{otherwise}
\end{cases}
\end{align}
where $T$ is a threshold level. For numerical stability in $\log$, we use $P_{t,\epsilon}(x) = \max(\epsilon,x)$ with a small positive number $\epsilon$ and $g^{inv}$ is the optimal inversion function minimizing the expected $\ell_2$ loss (See Appendix.~\ref{appen:noninvertible_rule}). 
For example, $g$ with $T=0.4$ is illustrated in Fig.~\ref{fig:g_nonlinear_noninv}. In our experiment, we used $T=0.02$, so any value of $m_3$ less than two percents will be ignored, and $\epsilon=10^{-10}$.

The averaged $\ell_2$ errors in percentage are, again for $N_{test}=10000, \sigma=0.005$,
\begin{align*}
e_{oracle} &= 0.872863 ,\\
e_{benchmark} &= 0.872274 ,\\
e_{ann} &= 0.403223 .
\end{align*}
Again, the direct comparison with the results coming from above other systems may not be feasible due to different system functions but the superiority of the ANN approach is evident. The bias introduced from the thresholding effect derived in Appendix.~\ref{appen:noninvertible_rule} is $ (0.02^3/12)^{0.5} \times  100\% = 0.082\%$ so the expected increased error is not large.

\subsubsection{Nonlinear systems: invertible transformation with an obfuscating variable}
\label{sssection:nonlinear_inv_obf}

We added an obfuscating variable to the invertible system described in the above Section \ref{sssection:nonlinear_inv}. The dimension of the unknowns became $M=4$. 

We assume that this obfuscating variable is not a dominant in that its weight is not greater than $20\%$. Generally, we can assume that the $\ell_1$ norm of the obfuscating variables are bounded. This is a reasonable assumption in practice too, because unknown variables outside our consideration or interest do not significantly determine the observations. If so, we would either include them in the model or research the physics to rebuild a model.

The errors from the oracle and benchmark estimators are calculated using equation \eqref{eq:ideal_loss_obs} where $\hat{\vv{m}} \in S^3$ contains only the scaled first 3 dimensional components such that $\sum_{i=1}^{M=3} \hat{m}_i = 1$.
\begin{align*}
e_{oracle} &= 1.066473 ,\\
e_{benchmark} &= 1.065830  ,\\
e_{ann} &= 0.557085 .
\end{align*}

In our experiment, we bound the obfuscating variable such that $m_4 \leq 0.2 $, which increases estimation error less than introducing a thresholding operation with the level 0.2 in one variable would because $(0.2^3/12)^{0.5} \times  100\% = 2.6\%$, and all the averaged errors above are less than $2\%$. In our simulation with $10000$ samples, the test error increase in the ANN approach is slightly less than those in the other approach in Section \ref{sssection:nonlinear_inv} but, this requires more investigation because their system functions are different with different input vectors.

\subsubsection{Nonlinear systems: noninvertible transformation with an obfuscating variable}
\label{sssection:nonlinear_noninv_obf}

We added an obfuscating variable to the noninvertible system described in Section \ref{sssection:nonlinear_noninv}. As in the previous experiment, $m_4 \leq 0.2 $ and the errors are increased compared to those in Section \ref{sssection:nonlinear_noninv}. 
\begin{align*}
e_{oracle} &= 1.089608 ,\\
e_{benchmark} &= 1.090022   ,\\
e_{ann} &= 0.534887 .
\end{align*}

\subsubsection{Nonlinear systems: transformation with varying magnitudes}
\label{sssection:nonlinear_H_scaled}

We define the following nonlinear system and experimented the ANN approach with $N_{train} =N_{test}=10000, M=3, L= 5, \sigma=0.005$ as in Section \ref{sssection:linear}.
\begin{align}
\label{eq:nonlinear_vary_mag}
\vv{s} = \| \mm{H} \vv{m} \|_2^2 \, \mm{H} \vv{m} + \vv{n}
\end{align}
This case cannot have oracle nor benchmar inversion results because we cannot estimate the scale factor $\| \mm{H} \vv{m} \|_2^2$ and the unknown variable $\vv{m}$ simultaneously without good prior knowledge. This inversion is called generally blind-deconvolution and semi-blind or myopic deconvolution with some prior knowledge of the unknown or the system \cite{Park2012}. 

\begin{figure} [h]
 \centering %  trim={<left> <lower> <right> <upper>}
 \includegraphics[width=4in,trim={0cm 0cm 0 0cm},clip]{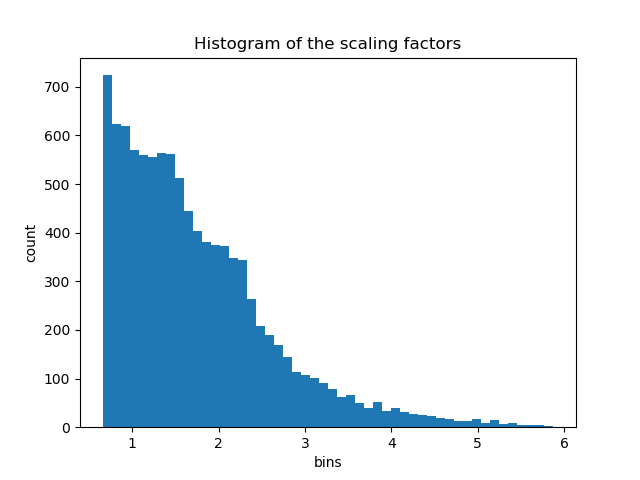}
 \caption{\label{fig:nonlinear_H_scaled_hist} Histogram of  $\| \mm{H} \vv{m} \|_2^2$, where majority ($79\%$) of the scaling factors are greater than 1, thus amplifying the signal compared to noise.
 }
\end{figure}

Our approach in this work estimates the inverse system in the ANN and the unknowns. The evaluated error shows the better result than other previous cases. 
\begin{align*}
e_{ann} &= 0.272344 .
\end{align*}
This better performance would be due to the effectively increased signal-to-noise ratio (SNR); the minimum of the scaling factors was $0.66$ and $79\%$ of the factors were larger than 1, as seen in Fig.~\ref{fig:nonlinear_H_scaled_hist}. The averaged $\ell_2$ norm $\| \mm{H} \vv{m} \|_2$ is $\E \| \mm{H} \vv{m} \|_2 = 1.28$ and $\E \| \mm{H} \vv{m} \|_2^2=1.74 $ with $\E$ being an empirical averaging operator here.

\subsubsection{Nonlinear systems: transformation with added correlations of unknowns}
\label{sssection:nonlinear_corr}

We designed another type of nonlinear system with a nonlinear function $g$ mapping from simplex to an auxilary vector \vv{z} below. 
\begin{align}
\vv{z} &= g(\vv{m}) = \transpose{[m_1 , 0.4 m_2, 0.2 m_1^2 , m_3^2 ,  m_1 m_2 ]}  , \\
\vv{s} &=  \mm{H} \vv{z} + \vv{n} .
\end{align} 
In this system response, the information of $m_1$ is abundant also with its original value, while $m_2, m_3$ are transformed and multiplied with others. We have more redundant intermediate variables of 5 dimensions from $\vv{m} \in S^3$ and the system matrix is enlarged, from $5\times 3$ to $5\times 5$, having more perturbations or variations in outputs. However, a large training set can accurately estimate the inverse system and the unknowns. Because the number of training samples seems large enough, the performance is similar to the linear case and other nonlinear cases as expected. 
\begin{align*}
e_{ann} &= 0.584665 .
\end{align*}
The oracle and benchmark cases are not evaluated because without knowing functional form or the intermediate dimension the estimators cannot be formulated. In contrast, the ANN approach is agnostic to such a knowledge of intermediate transformations and introduced correlations. If we assume this knowledge, then we can refer to the errorr levels in the linear system case in Section \ref{sssection:linear} and these should be comparable with the above ANN performance.

\subsubsection{Nonlinear systems: transformation with varying peak responses}
\label{sssection:nonlinear_peak_vary}

\begin{figure} [h]
 \centering %  trim={<left> <lower> <right> <upper>}
 \includegraphics[width=4in,trim={0cm 0cm 0 0cm},clip]{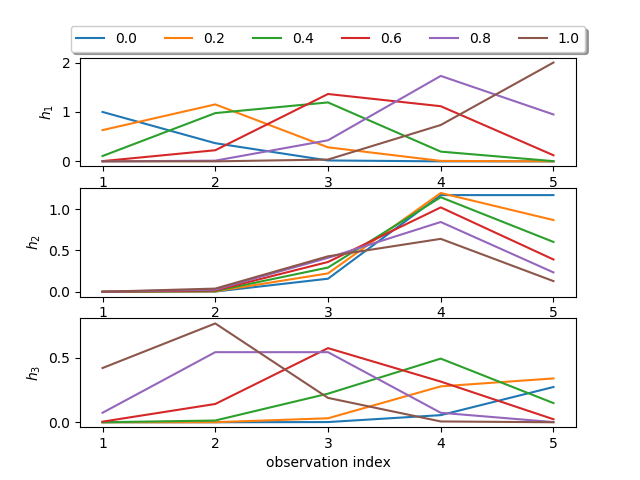}
 \caption{\label{fig:nonlinear_peak_vary} Varying responses in shape or peak locations of the component-wise system functions as its argument $m_i$ for $i=1,2,3$ changes, sampled at $m_i= [0,0.2,0.4,0.6,0.8,1]$. Different versions of $\vv{h}_i $s in Eq.~\ref{eq:nonlinear_peak_vary_h} are presented. $\vv{h}_1$ has a moving peak centered at the index 1 to 5 and the magnitude slightly increases as $m_1$ increases from 0 to 1, while $\vv{h}_3$ shows the opposite behavior in terms of the peak locations and magnitudes. $\vv{h}_2$ decreases slightly with shape changes as $m_2$ increases.
 }
\end{figure}

We define the following nonlinear system and experimented the ANN approach with $N_{train} =N_{test}=10000, M=3, L= 5, \sigma=0.005$ as in Section \ref{sssection:linear}.
\begin{align}
\label{eq:nonlinear_peak_vary}
\vv{s} &= \sum_{i=1}^M \vv{h}_i + \vv{n}, \\
\label{eq:nonlinear_peak_vary_h}
\vv{h}_i &= c_i \vv{g}_i ,\\
c_i &= (\mm{A}_{1,i} - \mm{A}_{2,i} ) m_i + \mm{A}_{2,i} ,\\
\vv{g}_i &= \exp \left[ \left( \vv{v} - \left((\mm{B}_{1,i} - \mm{B}_{2,i} ) m_i + \mm{B}_{2,i} \right) \right)^2 \right] ,
\end{align}
where $\vv{v}$ is an index vector $\vv{v} = \transpose{[1,2,\dots, L]}$ and 
\begin{equation}
\label{eq:A_B_matrices}
\mm{A}=\left[
\begin{array}{ccc}
2 & 0.7 & 0.8 \\
1 & 1.5 & 0.3 
\end{array}
\right], \,
\mm{B}=\left[
\begin{array}{ccc}
4 & 2.7 & 0.8 \\
0 & 3.5 & 4.3 
\end{array}
\right] .
\end{equation}
This system response has varying magnitudes dependent on composition weights $m_i$s in $c_i$s and different shapes also dependent on composition weights $m_i$s in $\vv{g}_i$s. Therefore, this case is more general than the one presented in Section \ref{sssection:nonlinear_H_scaled}. Fig.~\ref{fig:nonlinear_peak_vary} shows the varying responses in shape or peak locations of the component-wise system functions as its argument $m_i$ for $i=1,2,3$ changes, sampled at $m_i= [0,0.2,0.4,0.6,0.8,1]$. $\vv{h}_1$ has a moving peak centered at the index 1 to 5 and the magnitude slightly increases as $m_1$ increases from 0 to 1, while $\vv{h}_3$ shows the opposite behavior in terms of the peak locations and magnitudes. $\vv{h}_2$ decreases slightly with shape changes as $m_2$ increases.

The result shown below, from the ANN approach, is comparable with other cases but direct comparisons do not make much sense because the systems are different. 
\begin{align*}
e_{ann} &= 0.517819 .
\end{align*}
Again, the oracle and benchmark cases are not evaluated because it is difficult even with functional forms and parameter values due to complex nonlinearity. Instead, we provide the ratio of intensity, eg., $\ell_2$ norm, in noiseless system output of this system to that in linear system.
\begin{align}
\E \| \vv{z} \|_2 / \E \| \mm{H}\vv{m} \|_2 \approx 2 ,
\end{align}
where $\E$  is an empirical averaging operator here, $\vv{z}=\sum_{i=1}^M \vv{h}_i$ and \mm{H} is the same as used in Sections \ref{sssection:nonlinear_H_scaled} and \ref{sssection:linear}, $ \E \| {\mm{H}\vv{m}^{(t)}} \|_2 =1.28$ (reported also in Section \ref{sssection:nonlinear_H_scaled}) and $\E \| \vv{z} \|_2= 2.68$.
Considering only the amplified signal intensity we expect the better performance but the changing shapes must adversely affect the inversion performance.

\subsubsection{Nonlinear systems: transformation with varying peak responses wiht added correlations of unknowns}
\label{sssection:nonlinear_peak_vary_corr}

\begin{figure} [h]
 \centering %  trim={<left> <lower> <right> <upper>}
 \includegraphics[width=4in,trim={0cm 0cm 0 0cm},clip]{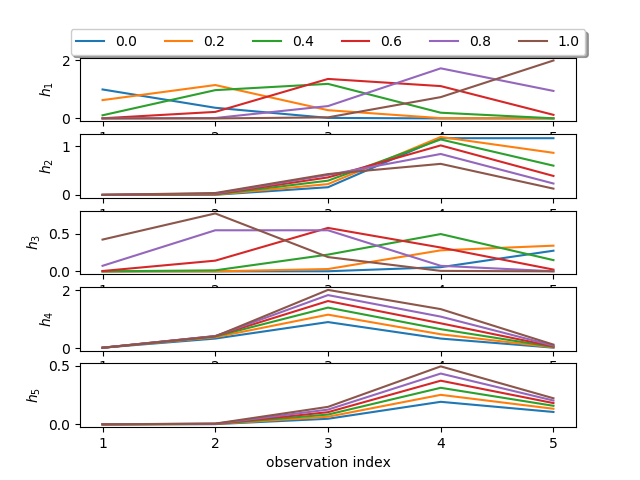}
 \caption{\label{fig:nonlinear_peak_vary_corr} Varying responses in shape or peak locations of the component-wise system functions as its argument $\tilde{m}_i$ for $i=1,2,3,4,5$ changes, sampled at $\tilde{m}_i= [0,0.2,0.4,0.6,0.8,1]$.  $\vv{h}_1,\vv{h}_2,\vv{h}_3$ are the same as in the previous sytem in Section \ref{sssection:nonlinear_peak_vary}  but with $\tilde{\vv{m}}$, a function of the unknown compositional vector $\vv{m}$. According to this function and the given system responses, a small quantity in $m_3$ seems difficult to estimate because its information is only in $\vv{h}_4, \vv{h}_5$ where small quantities of $\tilde{m}_i$ correspond to attenuated system responses. This would cause the degraded performance in inversion.
 }
\end{figure}

We define a similar nonlinear system to the previous system with $N_{train} =N_{test}=10000, M=3, L= 5, \sigma=0.005$ but with the added correlated terms.
\begin{align}
\vv{s} &= \sum_{i=1}^5 \vv{h}_i + \vv{n}, \\
\vv{h}_i &= c_i \vv{g}_i ,\\
c_i &= (\mm{A}_{1,i} - \mm{A}_{2,i} ) \tilde{m}_i + \mm{A}_{2,i} ,\\
\vv{g}_i &= \exp \left[ \left( \vv{v} - \left((\mm{B}_{1,i} - \mm{B}_{2,i} ) \tilde{m}_i + \mm{B}_{2,i} \right) \right)^2 \right] , \\
\tilde{\vv{m}} &= \transpose{[m_1, 0.4 \, m_2, 0.2 \, m_1^2, m_3^2 , m_2  m_3]},
\end{align}
where $\vv{v}$ is an index vector $\vv{v} = \transpose{[1,2,\dots, L]}$ and 
\begin{equation}
\label{eq:A_B_matrices}
\mm{A}=\left[
\begin{array}{ccccc}
2 & 0.7 & 0.8 & 2.2 & 0.5\\
1 & 1.5 & 0.3 & 0.9 & 0.2 
\end{array}
\right], \,
\mm{B}=\left[
\begin{array}{ccccc}
4 & 2.7 & 0.8 & 2.3 & 3.1 \\
0 & 3.5 & 4.3 & 2.0 & 3.2
\end{array}
\right] .
\end{equation}
Fig.~\ref{fig:nonlinear_peak_vary_corr} shows the varying responses in shape or peak locations of the component-wise system functions as its argument $\tilde{m}_i$ for $i=1,2,3,4,5$ changes, sampled at $\tilde{m}_i= [0,0.2,0.4,0.6,0.8,1]$. $\vv{h}_1,\vv{h}_2,\vv{h}_3$ are the same as in the previous sytem in Section \ref{sssection:nonlinear_peak_vary}  but with $\tilde{\vv{m}}$, a function of the unknown compositional vector $\vv{m}$. According to this function and the given system responses, a small quantity in $m_3$ seems difficult to estimate because its information is only in $\vv{h}_4, \vv{h}_5$ where small quantities of $\tilde{m}_i$ correspond to attenuated system responses. This would cause the degraded performance in inversion.
\begin{align*}
e_{ann} &= 0.723837 .
\end{align*}
Also, comparing the number to the previous system in Section \ref{sssection:nonlinear_peak_vary}, the added correlated terms did not help the inversion performance. Note that the direct comparison cannot be performed because $\vv{h}_2,\vv{h}_3$ are now linear and squared functions of $m_2,m_1$, not identity functions of $m_2,m_3$ as in the previous Section \ref{sssection:nonlinear_peak_vary}, resectively. 

\subsection{High dimensional linear systems}
\label{ssec:lin_highdim}
We experiment on high dimensional simplex variables.  
To simulate realistic experiments, we set  $M=20,L=1000$ to represent high dimensional spaces for the unknowns and observations.  We set  $ N_{test}=10000,  \sigma=0.005=.5\%$ and the designed system matrix in Fig.~\ref{fig:lin_highdim_H} has all nonnegative response curves. The designed system is given in Appendix.~\ref{app:lin_highdim_H}. For the training of the ANN, new $N_{train} = 10000$ samples every 100 epochs were generated to train the ANN because of the memory limitation while avoiding overfitting.  The samples in the training and tests sets are drawn according to the uniform distribution. The ANN is designed and tuned to the same parameter values as done in the previous experiments with linearly increased complexity of the networks as $M$ increases in the double layers of $4\times M$ nodes and another layer of $M$ nodes. 

\begin{figure} [h]
 \centering %  trim={<left> <lower> <right> <upper>}
 \includegraphics[width=5in,trim={0cm 0cm 0 0cm},clip]{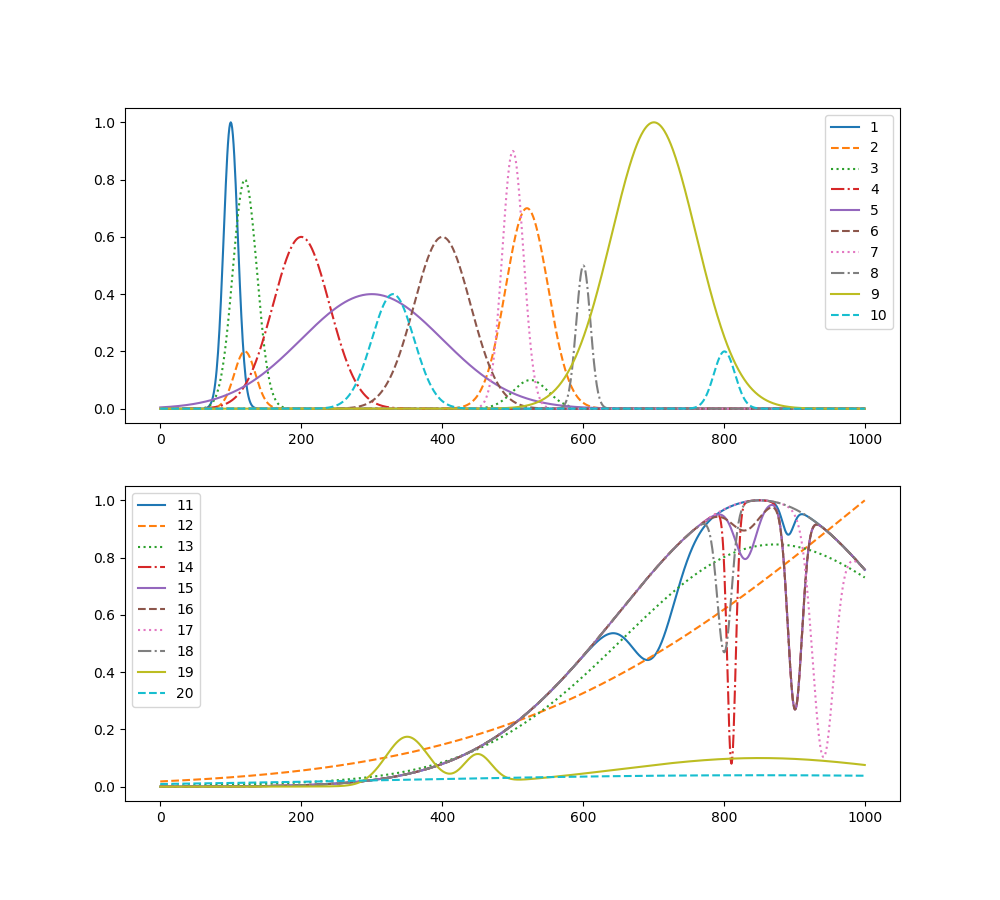}
 \caption{\label{fig:lin_highdim_H} Linear system matrix $\vv{H}$. Note that all response curves are nonnegative. The legends indicate the index $i$ in $\vv{h}_i$. The equations used to generated these curves are provided in Appendix.~\ref{app:lin_highdim_H}. The condition number of the matrix, the ratio of the largest singular value to the smallest, is 360.
 }
\end{figure}

\begin{figure} [h]
 \centering %  trim={<left> <lower> <right> <upper>}
 \includegraphics[width=4in,trim={0cm 0cm 0 0cm},clip]{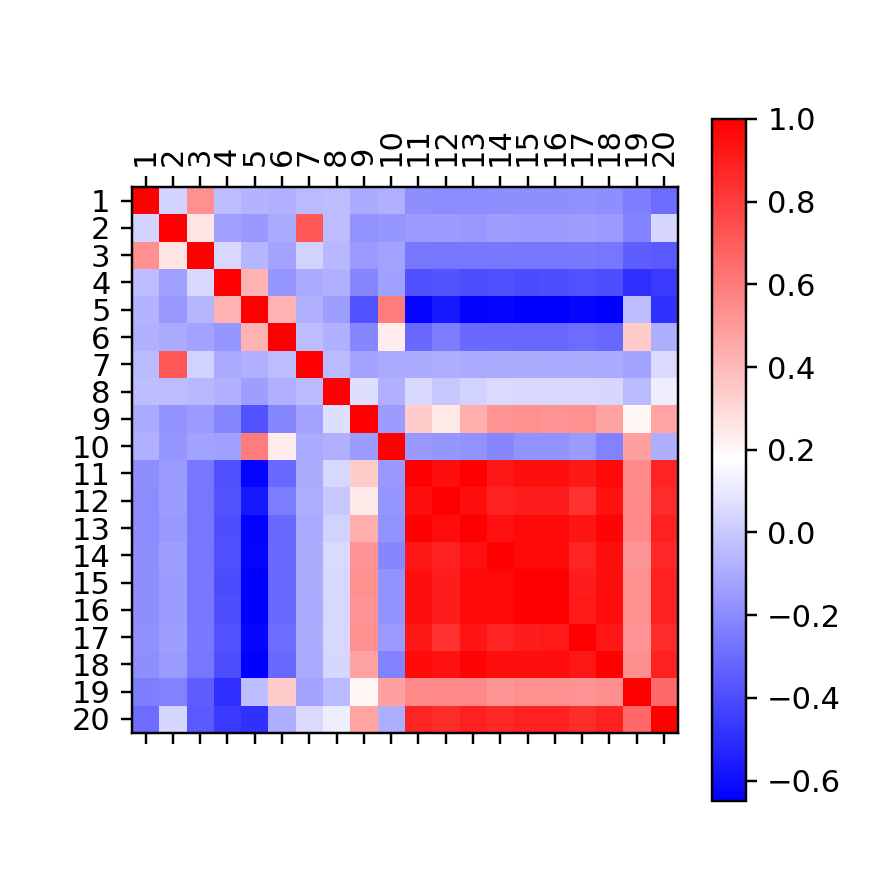}
 \caption{\label{fig:A_corr_01} Correlation of \mm{H} in Appendix.~\ref{app:lin_highdim_H}. As expected in Fig.~\ref{fig:lin_highdim_H}, the responses corresponding to the 11th through 20th compositions are similar, thus resulting in high correlation in red here. These high correlations cause difficulty in separation and reduction in estimation accuracy. 
 }
\end{figure}

From Fig.~\ref{fig:lin_highdim_H}, the correlations of components whose indices are 11 -- 20 must be significant because their overall envelop shapes are similar expect the valley shapes. These components have information residing in their valleys not envelope and the result high corelations are seen in the red block in Fig.~\ref{fig:A_corr_01}. Because of high correlations in the components number 11 -- 20, their estimation errors are higher than the components 1 -- 10, as seen in Fig.~\ref{fig:fig_nonlinear_cases_004}.

The trained system matrix for benchmark estimator is close to the true one because $$ \| \mm{H} -\hat{\mm{H}} \|_F / \| \mm{H}  \|_F = 0.0029386144131524146  . $$ The results on the test using the oracle and benchmark estimators are thus similar. 
\begin{align*}
e_{oracle} &= 3.779764 ,\\
e_{benchmark} &= 3.72622 ,\\
e_{ann} &= 2.214486. 
\end{align*}
The nonnegative high dimensional matrix with the larger condition number, 360, compared to that in low dimensional system, 3.23, degrades the performance from $0.57\%$ to more than $2\%$ errors. This can be seen visually in Fig.~\ref{fig:lin_highdim_H}, where there are many overlapped, similar shaped parts. However, the reported errors are still less than the theoretical bound for the unconstrained estimator \eqref{eq:oracle_est_error2}, 
$$ d_{oracle,uc} \times 100 =  4.7868905294620099 . $$ 

Moreover, the ANN approach outperforms the other two. 
Compared to the low dimensional linear case in Section.~\ref{sssection:linear}, the difference in the errors is significant. This must come from the locality of the ANN approach specific to the training set and the globality of the methods based on matrix pseudo-inversion. In the experiment, even with the uniform sampling in a simplex, the high dimensional simplex seems to exhibit locality with rare samples near the end-members ($100\% - \epsilon$) and relatively many samples away from them.

High dimensional simplex spaces may seem counter-intuitive particularly regarding the volume distribution. In fact, high dimensional simplices, along with other high dimensional polytopes, have the major volumn concentration on their surfaces but, near the corner, where the end-members are located, the volume diminishes as the dimension increases. This can be also demonstrated empirically by using uniform sampler on a simplex (see Appendix \ref{appen:thinVolCorner}). This implies that under the uniform distribution in a high dimensional simplex, the chance of drawing samples close to any end-members is negligible. 
However, in controlled experiments where observations are measured based on fabricated or designed samples on a simplex domain, as known as designed compositions, we can have the measurements corresponding to end-member compositions or pure contents of only one individual composition, i.e., $\vv{m}=\vv{e}_i$ for the $i$th end-member. Therefore, we can add the observations from end-members into our training set if we believe that the observations coming from near end-members are expected in practice.

To test the locality of the ANN and globality of the other two based on matrix inversion, 
we performed a simple test with the observations only from the $M$ end-members. Here, for the benchmark estimator, the training and test sets coincide on the $M$ observations, while the ANN estimator was already trained using the $N_{train}$ training samples. 
\begin{align*}
e_{oracle} &= 4.690122 ,\\
e_{benchmark} &\approx 0 ,\\
e_{ann} &= 30.940643   .
\end{align*}
The oracle estimator is indepedent of the training set and uses the true matrix, whose error is now much closer to but still less than $d_{oracle,uc}=4.79$, the benchmark uses the trained matrix and again use it for testing, leading to close to zero error as expected, and the ANN approach produces a significantly large error because there were extremely rare samples among $N_{train}=10000$ training samples that are close to any end-members.  Therefore, in practice if we believe there is a significant number of samples coming from near end-members, we should include them in the training data.

\subsection{High dimensional nonlinear systems}
\label{ssec:nonlin_highdim}

We defined a high dimensional nonlinear systems in Appendix.~\ref{app:highdim_nonlin}, where obfuscating variables and mixture models are also considered too. The system correlates some variables and transforms the original unknown vector with nonlinearly with fractional polynomials and exponential functions, thresholding, and shape changing with moving peaks and valleys. 
In this section, we experimented numerous ANN structures because of the higher order of complexity of the system: our base model with double layers of $4 \times M_{v}$, where $M_{v}$ is the number of components of interest or assumed, double layers of $16 \times M_v$, $32 \times M_v$, convolutional neural networks (CNN) of having a convolutional layer and then either double layers of $4 \times M_v$ or  $32 \times M_v$ feedforward networks.

\begin{figure} [h]
 \centering %  trim={<left> <lower> <right> <upper>}
 \includegraphics[width=5in,trim={0cm 0cm 0 0cm},clip]{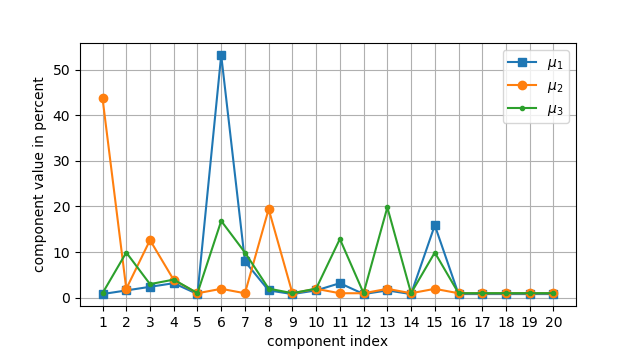}
 \caption{\label{fig:mixture_centers_01} Three centers in the mixture model, as defined in Eq.~\eqref{eq:mixture_centers}. $\mu_i$s is the center of the $i$th mixture and the remaining portion is filled with the samples drawn from the uniform distribution.
 }
\end{figure}

Additionally, we tested two cases for the compositonal distributions. One is the uniform distribution and the other is a mixture model.
In the designed mixture model, the mixture centers in percent are shown in Fig.~\ref{fig:mixture_centers_01}, 
and the corresponding $\sigma_i$s, the sample proportions, and details are provided in Appendix.~\ref{app:highdim_nonlin}. In the mixture model, there are still samples drawn from the uniform distribution. The drawn compositional vector can be truncated and normalized to satisfy the simplex condition. Also, in generating samples we disgard the samples whose $m_{19}, m_{20}$, as obfuscating variables, are greater than $5\%$. 
The result samples in $S^{20}$ with the described specification and corresponding noisy measurements using the nonlinear system with the observational noise level $0.005=0.5\%$ constitute the training and test sets.
In the experiments using the mixtures, we randomly shuffled the samples in training and test sets. We may retain the original compositional vector $\vv{m}^0 \in S^M (M=20)$ including obfuscating variables in $\vv{m}^1$, which is used to synthesize noisy measurments, but use $\vv{m} \in S^{M_v} ( {M_v}=18)$ without those variables for comparisons (Eq.~\ref{eq:ideal_loss_obs}). In other words, even if the responses of noisy observations embed the obfuscating variable effects, we do not use obfuscating varibles for training, and testing considers only the normalized version of the variables excluding obfuscating varibles. 

\begin{figure} [h]
 \centering %  trim={<left> <lower> <right> <upper>}
 \includegraphics[width=7in,trim={0cm 0cm 0 0cm},clip]{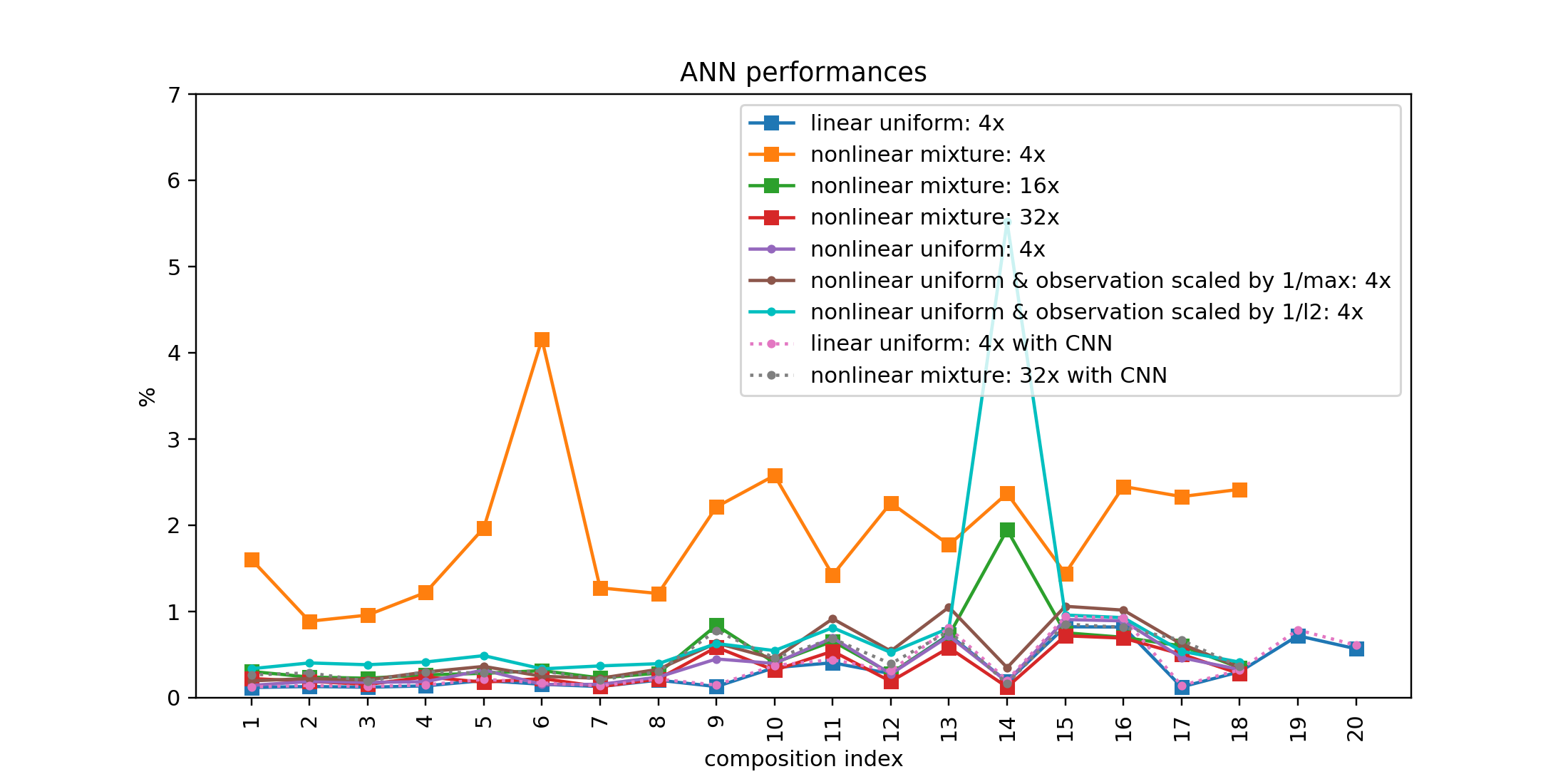}
 \caption{\label{fig:fig_nonlinear_cases_004} Componenent-wise error on the test set of high dimensionoal data, in average of absolute deviation in percent in y-axis. The legends correspond to the first 9 cases in orderin Table.~\ref{tab:errors}. 
 The simplest ANN structure applied in nonlinear systems may have under-fitting or convergence problem. Especially, component 14, corresonding to the signal of a moving peak, seems to cause the problem as the most difficult variable to estimate  especially in simpler models, while the models of the order of $32 \times M$ or larger do not exhibit such problems. 
 Two nonlinear systems have responses that are divided by their maxium or $\ell_2$ norm, resulting in added nonlinearity and slightly increased errors. For linear systems $M=M_v=20$ and for nonlinear systems $M =20, M_v = 18$ and there are two obfuscating variables. The CNN layers have a layer of 32 nodes and another of 16 with the kernel size 7, 3 strides, and ReLu activation. The component-wise $\ell_1$ error, the average of absolute deviation, is computed using Eq.~\eqref{eq:l1error}. The large model with $32\times M_v$ nodes and the linear cases show the low errors, as expected.
 }
\end{figure}

The performance in high dimensional examples with $L=1000$ are demonstrated by considering numerous cases of sample distribution, system types, and neural network structures. We added two nonlinear systems whose response is divided by its maxium or $\ell_2$ norm, resulting in added nonlinearity and slightly increased errors. 
We tried convolutional neural networks (CNN) too. We placed the convolutional layers before the double layers. The CNN layers consists of a layer of 32 nodes and another of 16 with the kernel size 7, 3 strides, and ReLu activation. 
For completeness, we included the results from linear systems in this section. For linear systems $M=M_v=20$ and for nonlinear systems $M =20, M_v = 18$ and there are two obfuscating variables. 
The $\ell_2$ error, as the overall error, is computed using Eq.~\eqref{eq:l2error} and reported in Table.~\ref{tab:errors}. 
The component-wise $\ell_1$ error, the average of absolute deviation, is computed using Eq.~\eqref{eq:l1error} and illustrated in Fig.~\ref{fig:fig_nonlinear_cases_004}.

We note that the two linear cases along with the largest model with double layers of $32 \times M$ or larger achieve the minimal errors due to the lowest complexity or the adaptive power, respectively.
Possibly, the two cases with double layers of $4\times M_v$ whose $\ell_2$ errors are more than 3 seem to have estimation bias or optimized insufficiently, because the optimization with the simple ANN structure showed too slow convergence emprically through many trials of different optimizers, tunings, and techniques.
In other words, the simplest ANN structure applied in nonlinear systems may have under-fitting or convergence problem in practice. Especially, component 14, corresonding to the signal of a moving peak, seems to cause the problem as the most difficult variable to estimate  especially in simpler models, while the models of the order of $32 \times M$ or larger do not exhibit such problems (Fig.~\ref{fig:fig_nonlinear_cases_004}). 
Generally,  increasing the number of nodes, $16$ or $32 \times M$ in our experiments, improves stability and accuracy without causing over-fitting by training on sufficient data. Adding a convolutional layer into our base structure with double layers of $\{4,16,32\}\times M$ may help but has not been extensively experimented in our work. 

\begin{table}
\begin{center}
\begin{tabular}{|l|l|l|l|l|}
\hline
System type & samples & ANN type or method & error \\ \hline\hline
linear      & uniform             & double layers of $4\times M_v$ & 2.21  \rule{0pt}{3ex} \\
nonlinear      & mixture          & double layers of $4\times M_v$ & 10.51 \rule{0pt}{3ex} \\
nonlinear      & mixture          & double layers of $16\times M_v$&  3.53 \rule{0pt}{3ex} \\
nonlinear      & mixture          & double layers of $32\times M_v$& 2.16 \rule{0pt}{3ex} \\
nonlinear      & uniform          & double layers of $4\times M_v$ & 2.36 \rule{0pt}{3ex} \\
nonlinear,  divided by its max            & uniform          & double layers of $4\times M_v$ & 2.98  \rule{0pt}{3ex} \\
nonlinear,  divided by its $\ell_2$ norm  & uniform          & double layers of $4\times M_v$ & 6.53  \rule{0pt}{3ex} \\
linear      & uniform             & CNN layers + double layers of $4\times M_v$ & 2.45  \rule{0pt}{3ex} \\
nonlinear      & mixture & CNN layers + double layers of $32\times M_v$ & 2.87  \rule{0pt}{3ex} \\ 
linear      & uniform             & pseudo-inverse of \mm{H} \& projection & 3.78  \rule{0pt}{3ex} \\
linear      & uniform & kNN (optimal k=11)  & 2.54  \rule{0pt}{3ex} \\ 
nonlinear      & mixture & kNN (optimal k=11)  & 6.54  \rule{0pt}{3ex} \\
nonlinear      & uniform & kNN (optimal k=11)  & 13.21  \rule{0pt}{3ex} \\
\hline
\end{tabular}
\end{center}
\caption[]{\label{tab:errors} Performance in high dimensional examples with $L=1000$. `samples' column indicates sample distributions in the simplex. Two nonlinear systems have responses that are divided by their maxium or $\ell_2$ norm, resulting in added nonlinearity and slightly increased errors. For linear systems $M=M_v=20$ and for nonlinear systems $M =20, M_v = 18$ and there are two obfuscating variables. The CNN layers have a layer of 32 nodes and another of 16 with the kernel size 7, 3 strides, and ReLu activation. The $\ell_2$ error is computed using Eq.~\eqref{eq:l2error}. We note that the two cases with double layers of $4\times M_v$ whose errors are more than 3 seem to have too slow convergence problems. The large model with $32\times M_v$ nodes and the linear cases show the low errors, as expected.
For comparison, we put the results in Section.~\ref{ssec:nonlin_highdim} using matrix inversion and projection and nearest neighbor interpolation methods described in Appendix.~\ref{app:kNN}. }
\label{tab:r}
\end{table}

For comparison, we put the results in Section.~\ref{ssec:nonlin_highdim} using matrix inversion and projection and k-nearest neighbor (kNN) interpolation methods described in Appendix.~\ref{app:kNN} in Table.~\ref{tab:errors}. 
For fair comparisons between ANN and kNN approaches, 
we use the same number of training samples; $N_{train} = 10000 , N_{test} = 1000 $. For stability of the kNN method, a computed distance is truncated to 0 when it is a negative number or $10^{10}$ when it is greater than $10^{10}$. 
For the linear system responses, the matrix inversion even with the known system matrix followed by the simplex projection is significantly inferior to ANN and kNN methods, while kNN method with optimal setting ($k=11$) can compete with ANN approaches. 
However, for the nonlinear system, kNN produced large biases. For simplicity of the figure, we did not include component-wise errors in Fig.~\ref{fig:fig_nonlinear_cases_004}. Another drawback in using kNN estimators is the increasing computation time in application or testing as the training set increases. This is because a test sample needs to be compared to the whole training set. This drawback can be mitigated by using tree building and searching algorithms but is out of scope of our work. 
One interesting observation from kNN approach is that it performs better in interpolating the concentrated samples as in the designed mixture than in exptrapolating between the scattered samples as in the uniformly distributed samples.

\section{Conclusion}

We demonstrated the promising performances in estimating the compositional unknown vectors using our simple ANN design throughout our extensive experiments. 
The ANN approaches can compete with the optimal bounds for linear systems, where efficient convex optimization theory applies and there are guranteed global optima. However, in complex nonlinear system inversions, we do not have such benchmark or global properties. We thus provided several surrogate bounds or analysis and performed extensive experiments by designing numerous different types of nonlinear systems, both in low and high dimensions.
In our experiments with low noise level, we demonstrated that the double layers of $4 \times M_v$ through $4 \times (LM_v)^{0.5}$ nodes in ANNs guarantee good estimation performances. We thus conjecture that the double layers of such order are sufficient in other nonlinear systems with other noise levels and will work on this as our future work.

The estimation performance may depend on the desired distribution in the data. We simulated mostly uniform distribution on simplices because it is the most scattered distribution as the worst case in the volume measure of a simplex set. We also performed additional experiments using mixtures of concentrated distributions and uniform distribution.

It is worth to note that the uniform distribution on high dimensional simplex shows counter-intuitive characteristics in terms of rare chance of selecting near any end-members. In this sense, the drawn samples seem to exhibit concentrations away from any component being large because of the low probability of selecting near end-members whose major component is $(100\% - \epsilon)$ with a small positive number $\epsilon$. Indeed, the probability decreases exponentially as the dimension increases. 
As was done in our experiments with mixtures, we can include samples concentrated near end-members into the training set for estimation of such compositions. 

Even though the considered nonlinear system types in this paper may not cover all the possible system types, this work covers numerous different types of nonlinear systems by our designing them and extensive experiments. An investigation of other possible types is our future work.

Another future work is to find the minimum or optimal depth of ANNs to effectively invert a nonlinear system that can be perfectly represented with terms up to order $\alpha$ as in Taylor series approximations of a nonlinear system in Appendix.~\ref{appen:TaylorExpansion}. This work would further embolden our empirical conclusion that the ANNs with nonlinear activations are sufficient to provide good inversions. We conjecture that the order of $\alpha +1 $ layers would be sufficient according numerous experiments we tried.

%%%%%%%%%%%%%%%%%%%%%%%%%%%%%%%%%%%%%%%%%%%%%%%%%%%%%%%%%%%%%%%%%%%%%%%%%%%%%%
\appendix
%\section*{Appendices}
%%%%%%%%%%%%%%%%%%%%%%%%%%%%%%%%%%%%%%%%%%%%%%%%%%%%%%%%%%%%%%%%%%%%%%%%%%%%%%

\subsection{Nonuniqueness of parameters of a shallow network for sum-constant vectors}
\label{app:nonuniqueness_shallow}

The parameters in a shallow network $\mm{W},\vv{m}$ are not uniquely determined in learning \mm{A} using the training set $\{\vv{x}_i,\vv{y}_i\}_i$ coming from $\vv{y} = \mm{A} \vv{x}$.
Let $\transpose{\vv{1}} \vv{x} = K$ and $\mm{W} = \mm{A} - \vv{b}\transpose{\vv{1}} / K$ with an arbitrary constant vector \vv{b}. But, the ANN output always matches $\vv{y}$.
\begin{equation}
\mm{W}\vv{x} + \vv{b} = \mm{A}\vv{x} - \vv{b} \sum x_i / K = \mm{A}\vv{x} - \vv{b} = \vv{y} 
\end{equation}

\subsection{Optimal inversion estimator for a partially noninvertible thresholding operator}
\label{appen:noninvertible_rule}
Let a random variable $x \in \mathbb{R}^1$ to be estimated follow a uniform distribution on its domain $[0, U_x]$. 
We assume that a partially noninvertible thresholding operator, such as hard or soft-thresholding, has a noninvertible region on $[0,T]$ with $T<U_x$. Except the thresholded region, a perfect inversion is achieved so the estimator $\hat{x} = x$ on $[T,U_x]$. Without loss of generality, we let $U_x = 1$ for a simple derivation.

We can obtain the best estimator minimizing $\ell_2$ distance between $x$ and $\hat{x}$.
The squared loss function is then defined with the expectation with respect to the probability function $P(x)$.
\begin{equation}
L_{threshold}^2 = \E \| x - \hat{x} \|_2^2 = \int_0^T (x - \hat{x})^2 d P(x) = T (\hat{x} - T/2)^2 + T^3/12 
\end{equation}
Therefore, the minimum $L_{threshold}^*$ is achieved when $\hat{x} = T/2$
\begin{equation}
L_{threshold}^* = \sqrt{T^3/12 }
\end{equation}
Note that the maximum error is $| x -\hat{x}| = T/2$

In practice, using Monte Carlo simulations, 
$L_{threshold}^2 \approx \| \vv{x} - \hat{\vv{x}} \|^2_2 / N  $ with $\vv{x}\in [0, U_x]^N$, leading to 
\begin{equation}
L_{threshold} \approx \| \vv{x} - \hat{\vv{x}} \|_2 / \sqrt{N} 
\end{equation}

For example, in a simplex domain, when there is a thresholding on $[0,0.1]$, the best estimator, assuming uniform distribution of the unknown, will predict $0.05$ on any input in the region. The maximum error $| x -\hat{x}|$ is $0.05$, but the overall loss $L_{threshold} = 0.00913$ or near 1\% error.

\subsection{Taylor expansion on a simplex}
\label{appen:TaylorExpansion}
A general model with a differentiable $h$ can be practically decomposed and well-approximated with Taylor expansion.  Without loss of generality, let we consider the series centered at $\vv{0}$. A general system response 
\begin{align}
h(\vv{m}) &\approx \sum_{k}  S_k \\
  S_k &= \sum_{i_1} \cdots \sum_{i_k} \vv{h}_{i_1,...,i_k} \,  m_{i_1} \cdots m_{i_k}
\end{align}
where $\vv{h}_{i_1,...,i_k}$ is a derivative with respect to $m_{i_1} \cdots m_{i_k}$ and $i_j \in [1,2, ..., M]$ for $ \forall j$.

\subsection{Volumne concentration in a high dimensional simplex}

\subsubsection{Thin concentration of volume in a self-similar corner of a high dimensional simplex}
\label{appen:thinVolCorner}

Let $V_M$ be the volume of a polytope in $M$ dimension without degenerate dimensions such as in a simplex. Then the volume of a self-similar polytope, whose size is $\epsilon < 1 $ times smaller, is $V_M \epsilon^M$. The volumne of this smaller polytope decreases as the dimension increases with the exact rate of $\epsilon^M$. This polytope can be placed to cover a corner inside the original polytope if it is convex. 

Specifically, the volumne of M-simplex $S^M$, 
according to \cite{Park2019spe},  is 
\begin{equation}
V(S^M) = \frac{\sqrt{M}}{ (M-1)!}
\end{equation}

Without loss of generality, considering the first axis of $S^M$, we define the subset $S^M_{\epsilon,1}$ located around the corner of the first which is the endmember $\vv{e}=\transpose{[1,0,\cdots,0]}$
\begin{equation}
S^M_{\epsilon, 1} =\{  (x_1, \ldots , x_M) \in S^M \, : \,   1-\epsilon \leq x_1 \leq 1  \}  .
\end{equation}
The volume of the self-similar $\epsilon$-sized polytope $S^M_{\epsilon, 1}$ is
\begin{equation}
V(S^M_{\epsilon,1}) = V(S^M) \epsilon^{M-1} ,
\end{equation}
with the dimensional factor $M-1$, due to one degenerate dimension of a simplex. 
The volume ratio 
$V(S^M_{\epsilon,1}) / V(S^M) = \epsilon^{M-1} \rightarrow 0$ as $M$ increases. Therefore, the contribution of the volume of any corner of a vertex diminishes in high dimensions because $M \epsilon^{M-1} \rightarrow 0$ as $M$ grows.

This behavior of a unit simplex is different in high dimensions from other polytopes such as a unit hyper cube. 
Note that a unit hyper cube in the dimension $M$ denoed by $C^M = [0,1]^M$ has volume of 1. Consider the set of the thin ($\epsilon$-thick) slice of the cube covering the first coordinate value of 1:
\begin{equation}
C^M_{\epsilon,1} =\{  (x_1, \ldots , x_M) \in [0,1]^M \, : \,   1-\epsilon \leq x_1 \leq 1  \}  
\end{equation}
and the volume of it is constant and not decaying over dimension.
\begin{equation}
V(C^M_{\epsilon,1}) = \epsilon 
\end{equation}

Therefore, the realization of a uniform random variable on a simplex in a high dimension produces near end members more rarely as the dimension grows; 
under a uniform distribution in a simplex $S^M$, an arbitrary volume inside the simplex is proportional to the probability that a drawn sample of the uniform random variable is within the volume. Therefore, the probability of a drawn sample being within $\epsilon$ distance from the $i$th end-member, $x_i=1$, is $P(x_i \in [1-\epsilon, 1] ) = \epsilon^{M-1}$ for a sample $\vv{x} \in S^M$. 
When $\epsilon > 0.5, M \geq 2$, because there are $M$ vertices, in $S^M$ the probability of a sample being within $\epsilon$ distance from any end-members is 
\begin{align}
\label{eq:P_epsilon}
P_\epsilon = P(x_i \in [1-\epsilon, 1] \mbox{ for } \forall i) = M \epsilon^{M-1}
\end{align}

Empirically, we can verify this exponential decrease in the number of drawn samples according to the uniform distribution, by evaluating the ratio of the number of samples, one of whose entries is over a specified number $1-\epsilon$, to the number of total drawn samples. 
With $N=10^6$ samples drawn according to the uniform distribution, we set several values of $\epsilon := 1-T$ to observe samples whose first component is greater than $T$. The result of this experiment is presented in Fig.~\ref{fig:P_thr_MC}.
\begin{figure} [h]
 \centering %  trim={<left> <lower> <right> <upper>}
 \includegraphics[width=4in,trim={0cm 0cm 0 0cm},clip]{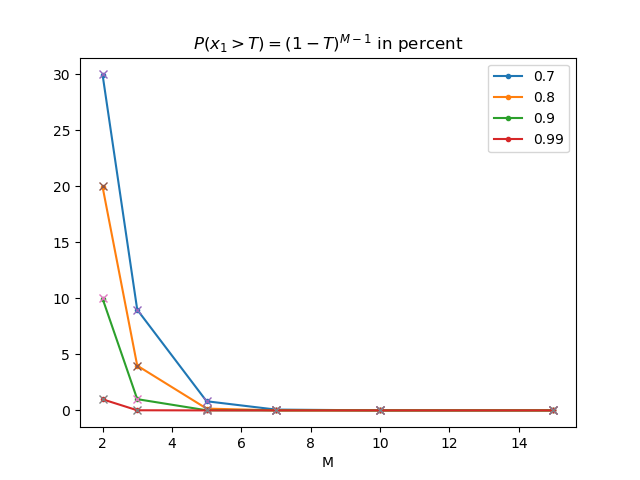}
 \caption{\label{fig:P_thr_MC} Sample ratios of $\vv{m} > T$ with $T = 0.7, 0.8, 0.9, 0.99$, $\vv{m} \in S^M, M = 2,3,5,7,10,15$. For many cases there was no drawn sample among one million samples, indicating 0. This demonstrates that the near endmember samples are rarely drawn in high dimensional simplices.
 }
\end{figure}

\begin{figure} [h]
 \centering %  trim={<left> <lower> <right> <upper>}
 \includegraphics[width=4in,trim={0cm 0cm 0 0cm},clip]{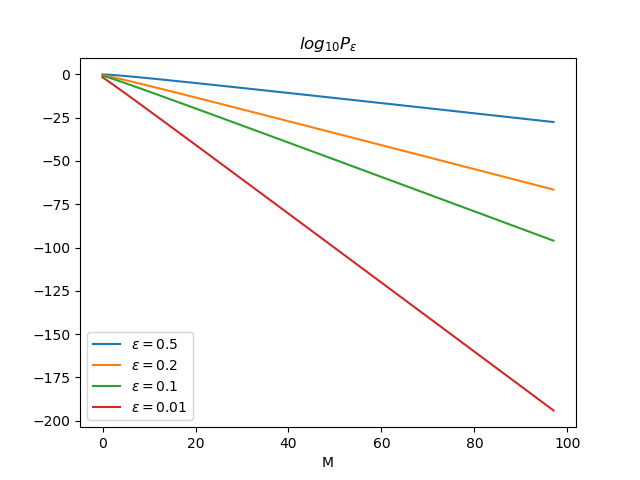}
 \caption{\label{fig:P_thr_MC} $P_\epsilon$ in Eq.~\ref{eq:P_epsilon} as the volume or probability of a sample having any component within $\epsilon$ distance within any end-members. The observed exponential decay with the log scale y-axis states that samples are extremely rare near any end-members under the uniform distribution.
 }
\end{figure}

\subsubsection{Still thin concentration of volume above the center of a high dimensional simplex}

We present another behavior of sampling related with the mean of the uniform distribution in simplices or a center of simplices. 
The mean of uniform distribution on simplex $S^M$ \cite{Park2019spe} is $\mu_M = \vv{1}/M$ and the probability of a drawn sample having one entry, e.g., $x_1$, greater than $c/M$ ($c$ time than the mean) is 
\begin{align}
\label{eq:P_cmu}
P(x_1 \in (c\mu_M, 1]  ) = P(x_1 > c\mu_M )= (1-c/M)^{M-1} 
\end{align}
This bound converges as the following 
\begin{align}
\label{eq:P_cmu_asym}
P(x_1 > c\mu_M ) \rightarrow e^{-c}, \mbox{ as } M \rightarrow \infty 
\end{align}
The equivalent $\epsilon $ value in \eqref{eq:P_epsilon} is $1-c/M$ and 
\begin{align}
M P(x_1 > c\mu_M ) \geq P(x_i \in [1/M, 1] \mbox{ for } \forall i)  \approx M e^{-c},
\end{align}
with large $M$ and $c$.
 
\begin{figure} [h]
 \centering %  trim={<left> <lower> <right> <upper>}
 \includegraphics[width=4in,trim={0cm 0cm 0 0cm},clip]{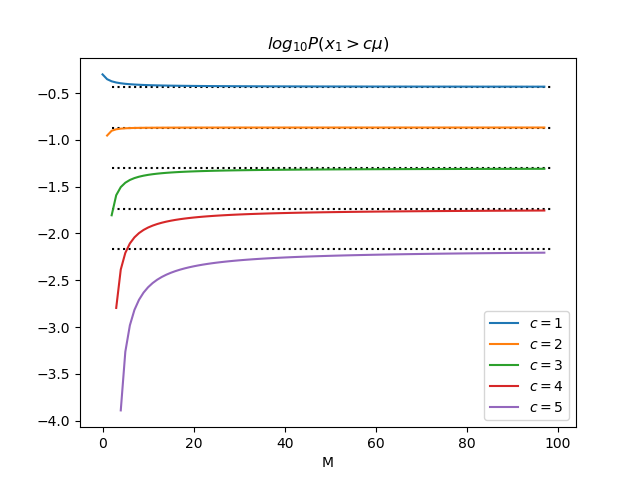}
 \caption{\label{fig:P_x1_cmu} $P(x_1>c\mu)$ of Eq.~\eqref{eq:P_cmu} with the asymptotes of Eq.~\eqref{eq:P_cmu_asym} in dotted lines in the log scale y-axis.
 }
\end{figure}

\begin{figure} [h]
 \centering %  trim={<left> <lower> <right> <upper>}
 \includegraphics[width=4in,trim={0cm 0cm 0 0cm},clip]{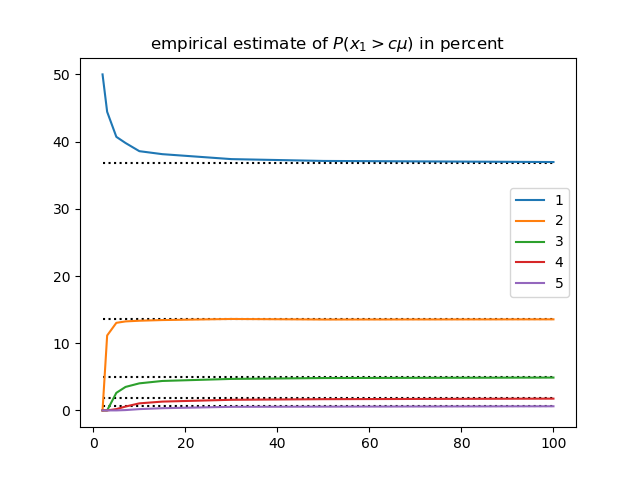}
 \caption{\label{fig:P_x1_cmu_MC} Monte Carlo estimates of Eq.~\eqref{eq:P_cmu} using 10,000 samples with the asymptotes of Eq.~\eqref{eq:P_cmu_asym} in dotted lines, along $M$ in percent in y-axis. 
 }
\end{figure}

Several curves of the probability $P(x_1>c\mu)$ of Eq.~\eqref{eq:P_cmu} with the asymptotes of Eq.~\eqref{eq:P_cmu_asym} are presented in Fig.~\ref{fig:P_x1_cmu} with the theoretical values and in Fig.~\ref{fig:P_x1_cmu_MC}  with Monte Carlo estimates using 10,000 samples.


\subsubsection{High concentration of volume near the center in a high dimensional simplex}

The probability of a sample lying in a band centered in the mean, under the uniform distribution, is
\begin{align}
\label{eq:P_band}
P( | x_1 -\mu | \geq a ) &\leq \frac{Var(x_1)}{a^2} \\
\label{eq:P_band2}
&= \frac{1}{a^2} \frac{M-1}{M+1} \frac{1}{M^2}   \\
\label{eq:P_band4}
&= \frac{1}{k^2} \frac{M-1}{M+1}  \mbox{ with $a=k/M$ } \\
\label{eq:P_band5}
&\rightarrow \frac{1}{k^2} \mbox{ as $M$ increases} ,
\end{align}
where the variance of a first component $Var(x_1)$ is computed using the first and second moments in \cite{Park2019spe}. 
For a fixed bandwidth $2a$, Eq.~\eqref{eq:P_band2} states that the chance of selecting a sample drawn outside the band $[\mu - a, \mu+a]$ decreases with the asymptotic rate of $1/M^2$. For the band $[ \mu-k/M , \mu+k/M]$, which linearly decreases as the dimension $M$ increases, Eq.~\eqref{eq:P_band4} states that the chance of selecting a sample drawn outside it is asymptotically constant. This analysis reveals the concentration near the mean in a high dimensional simplex in terms of its volume. This is comparable to the volume concentration near the surface or boundary in high dimensional cubes or spheres, because the above mean $\mu$ or the volume center in a simplex is close to its boundary in a high dimension.

\subsection{Equations used to generate the linear system matrix in Section \ref{ssec:lin_highdim} }
\label{app:lin_highdim_H}
We first define a radial basis function $\phi$.
\begin{align}
\phi(a,b) = \exp\left( -\dfrac{ (\vv{v} - a)^2} {2 b^2}   \right), 
\end{align}
where $a,b$ are real numbers and $\vv{v}$ is an index vector $\vv{v} = \transpose{[1,2,\dots, L]}$ and $\vv{v} - a := \vv{v} - a \vv{1} $. We define $\vv{h}_i$ as follows:
{\scriptsize
\begin{align*}
\vv{h}_1 &= \phi(100,10) \\
\vv{h}_2 &= 0.2 \phi(120,15) + 0.7 \phi(520,30) \\
\vv{h}_3 &= 0.8 \phi(120,17) + 0.1 \phi(525,25)  \\
\vv{h}_4 &= 0.6 \phi(200,40) \\
\vv{h}_5 &= 0.4 \phi(300,100) \\
\vv{h}_6 &= 0.6 \phi(400,40) \\
\vv{h}_7 &= 0.9 \phi(500,15) \\
\vv{h}_8 &= 0.5 \phi(600,10) \\
\vv{h}_9 &= \phi(700,60) \\
\vv{h}_{10} &= 0.2 \phi(800,15) + 0.4 \phi(330,30)  \\
\vv{h}_{11} &= \phi(850,200) - 0.3 \phi(700,30) - 0.1\phi(890,8) \\
\vv{h}_{12} &= 3 \phi(1500,500) / \max(3 \phi(1500,500))\\
\vv{h}_{13} &= 0.7 \phi(850,200) + 0.2 \phi(1500,500) / \max( \phi(1500,500)) \\
\vv{h}_{14} &= \phi(850,200) - 0.7 \phi(900,10) - 0.9 \phi(810,6)\\
\vv{h}_{15} &= \phi(850,200) - 0.7 \phi(900,10) - 0.2 \phi(830,15) \\
\vv{h}_{16} &= \phi(850,200) - 0.7 \phi(900,10) - 0.1 \phi(830,20) \\
\vv{h}_{17} &= \phi(850,200) - 0.8 \phi(940,15) \\
\vv{h}_{18} &= \phi(850,200) - 0.5 \phi(800,10) \\
\vv{h}_{19} &= 0.1 \phi(850,200) + 0.17 \phi(350,30) + 0.1 \phi(450,20)\\
\vv{h}_{20} &= 0.04 \phi(850,500) . 
\end{align*}
}

\subsection{An example of high dimensional nonlinear systems}
\label{app:highdim_nonlin}

A soft-thresholding function is defined as the following
\begin{align}
\label{eq:highdim_nonline_f_T}
f_T(\vv{m}) = \max(\vv{m} - T \vv{1}, \vv{0}),
\end{align}
with $T=0.03=3\%$.

A function that produces an extended version with correlated terms is defined as the following
\begin{align}
\label{eq:highdim_nonline_g}
g(\vv{x}) = \transpose{[x_1,x_2,x_3,x_1 x_2, 3 x_2 x_3]} \in \mathbb{R}^5,
\end{align}
with $\vv{x} \in \mathbb{R}_+$ (non-negative real set).

The system function is designed as the following
\begin{align}
\vv{h} &= h(\vv{m}) = \mm{H} \vv{z}  + \vv{y}_{4} +  \vv{y}_{14} + \vv{y}_{21} + \vv{y}_{22}  \\
z_k &= m_k \mbox{ for } k = 1,2,3,8,{10}, 12,13, 15,16,19,20 \\
z_k &= 0 \mbox{ for } k = 4,14 \\ 
[z_5,z_6,z_7] &= [g_1,g_2,g_3] = [m_5,m_6,m_7]\\ 
\vv{g} &= g(\transpose{[m_5,m_6,m_7]}) =: \transpose{[g_1,g_2,g_3,g_4,g_5]} \\ 
z_9 &= f_T(m_9) \\
z_{11} &= \exp( f_T(m_{11})) - 1 \\
z_{17} &= m_{17}^{1.5} \\
z_{18} &= m_{18}^{0.9} + m_{18}^2 \\
\vv{y}_{4} &=  0.6 \phi(peak_4,40) \times  m_4\\
peak_4 &= 100 m_4 + 200 \mbox{ (moving peak) } \\
\vv{y}_{14} &=  \left( \phi(850,200) - 0.7 \phi(valley_{14},10) - 0.9 \phi(810,6) \right) \times m_{14}\\
valley_{14} &= 100 (1-m_{14}) + 820 \mbox{ (moving valley) } \\
\vv{y}_{21} &= \phi(350,130) g_4 = \phi(350,130) m_5 m_6 \\
\vv{y}_{22} &= \phi(450,70) g_5 = \phi(450,70) m_6 m_7 ^3 
\end{align}
with  $f_T(\vv{m})$ in Eq.~\eqref{eq:highdim_nonline_f_T}, $g(\vv{x})$ in Eq.~\eqref{eq:highdim_nonline_g}, \mm{H} in Appendix.~\ref{app:lin_highdim_H}, and $\vv{m} \in S^{20}$, which leads to $\vv{z} \in \mathbb{R}_+^{20}$.

For the $i$th mixture having the sample proportion of $p_i$, the samples are drawn according to Gaussian distribution with mean $\vv{\mu}_i$ and the covariance $\sigma_i \mm{I}$. 
The mixture centers in percent are defined as the followings also shown in Fig.~\ref{fig:mixture_centers_01} 
{\scriptsize
\begin{align}
\label{eq:mixture_centers}
\vv{\mu}_1 &=      \transpose{      [0.79, 1.59, 2.38, 3.17, 0.79, 53.17, 7.94, 1.59, 0.79, 1.59, 3.17, 0.79, 1.59, 0.79, 15.87, 0.79, 0.79, 0.79, 0.79, 0.79] } \\
    \vv{\mu}_2&=   \transpose{     [43.69, 1.94, 12.62, 3.88, 0.97, 1.94, 0.97, 19.42, 0.97, 1.94, 0.97, 0.97, 1.94, 0.97, 1.94, 0.97, 0.97, 0.97, 0.97, 0.97] } \\
    \vv{\mu}_3&=   \transpose{     [0.99, 9.90, 2.97, 3.96, 0.99, 16.83, 9.90, 1.98, 0.99, 1.98, 12.87, 0.99, 19.80, 0.99, 9.90, 0.99, 0.99, 0.99, 0.99, 0.99] } 
\end{align}
}
and the corresponding $\sigma_i$s  in percent are
\begin{align}
\label{eq:mixture_sigmas}
\sigma_1 =  1 \% , \, %\\
    \sigma_2 = 2 \% , \,% \\
    \sigma_3 = 3 \%
\end{align}
with the sample proportions 
\begin{align}
\label{eq:mixture_proportions}
p_1 =  0.2 , \,%\\
p_2= 0.2  , \,%\\
p_3= 0.3 
\end{align}
while the remaining proportion of 0.3 ($=1-0.2-0.2-0.3$) is filled with the samples drawn according to the uniform distribution. Moreover, to satisfy the simplex condition on the samples coming from the mixtures, the negative components are truncated to zero and the components greater than one are truncated to one, followed by the scaling or normalization step with the $\ell_1$ norm of the possibly truncated vector. 
Note that the truncation can lead to the scaling of the vector for normalization and the final sample distribution can be non-Gaussian. However, because this truncation rarely occurs from our experiments with low $\sigma_i$s, the result distribution is approximately Gaussian.
After this, we disgard the sample whose $m_{19}, m_{20}$, as obfuscating variables, are greater than $5\%$. 
If we want to include end-members in the training or test set, we generate them except the end-members of obfuscating variables.
Now, let $G$ be such a sample generator function that generates the samples following the uniform distribution or a mixture in $S^{20}$ with the selected specification described above and corresponding noisy measurements with noise level $0.005$. 
In experiments, when the mixture model is used, we randomly shuffled the samples in training and test sets.
We may retain the original compositional vector $\vv{m}^0$ including obfuscating variables in $\vv{m}^1$, which is used to synthesize noisy measurments, but use $\vv{m}$ without those variables for comparisons (Eq.~\ref{eq:ideal_loss_obs}).

\subsection{Nearest neighbors estimators }
\label{app:kNN}
We provide the details of $k$-nearest neighbors (kNN) estimators that we use for comparisons.
From the test observation vector \vv{y}, we evaluate the distance $d(\vv{y}_j,\vv{y})$, which the squared Euclidean distance between \vv{y} and $j$th observation vector, $\vv{y}_j$, in the training set. Let $I_k(\vv{y})$ be the index set of the training samples having the $k$ smallest distances to \vv{y}. 
\begin{align}
\hat{\vv{x}} = \sum_{j \in I_k(\vv{y})} l_j  \vv{x}_j \, /  \sum_{j \in I_k(\vv{y})} l_j , 
\end{align}
where $l_j = 1/d(\vv{y}_j,\vv{y})$ and $d(\vv{y}_j,\vv{y}) = \|\vv{y}_j - \vv{y} \|_2^2$.
As a special case of $k=1$, the estimate will be the closest neighbor in the train set.

The performance of kNN depends on the number of neighbors $k$ and the distribution of the data. 
When $k$ is too small, the estimator does not use enough neighbors information. When it is too large, the estimator averages too many neighbors and insensitive to the given sample. Therefore, the optimal $k$ under a given data distribution should be an intermediate number. The detailed theoretical analysis of kNN performance is out of scope of this work but we empirically demonstrate this by evaluating the performance for our designed distribution of samples and corresponding system outputs. 

For fair comparisons with ANN approaches, 
we set $N_{train} = 10000 , N_{test} = 1000 $. For stability of the method, a computed distance is truncated to 0 when it is a negative number or $10^{10}$ when it is greater than $10^{10}$. 

\subsubsection{kNN in a linear low dimension}
\label{app:kNN_lin_lowdim}

\begin{figure} [ht]
 \centering %  trim={<left> <lower> <right> <upper>}
 \includegraphics[width=4in,trim={0cm 0cm 0 0cm},clip]{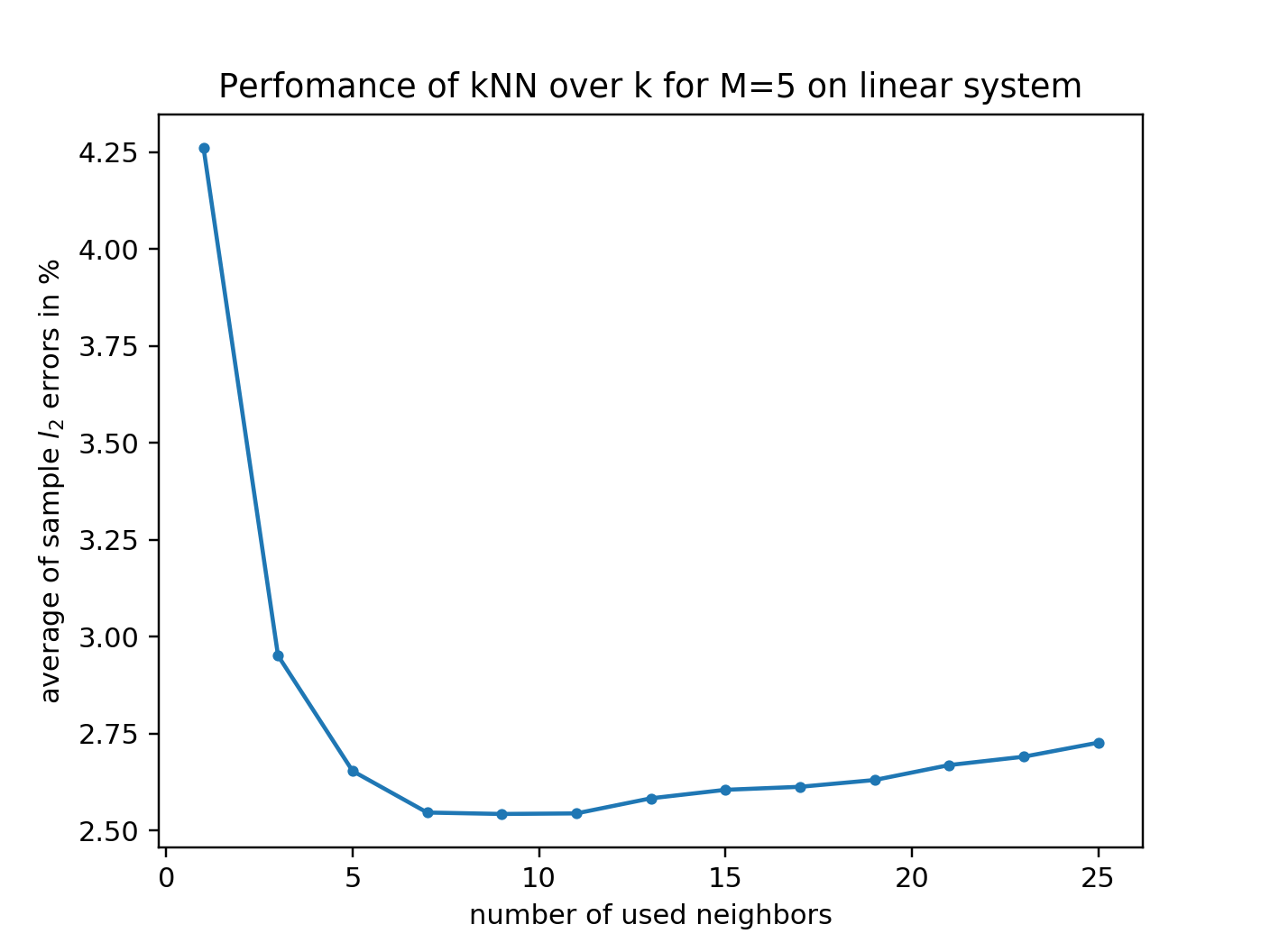}
 \caption{\label{fig:kNN_performance_01} The average of $\ell_2$ errors in percent (Eq.~\eqref{eq:l2error}) as the performance of kNN in a linear low dimensional system with $L=7, M=5$. }
 \includegraphics[width=4in,trim={0cm 0cm 0 0cm},clip]{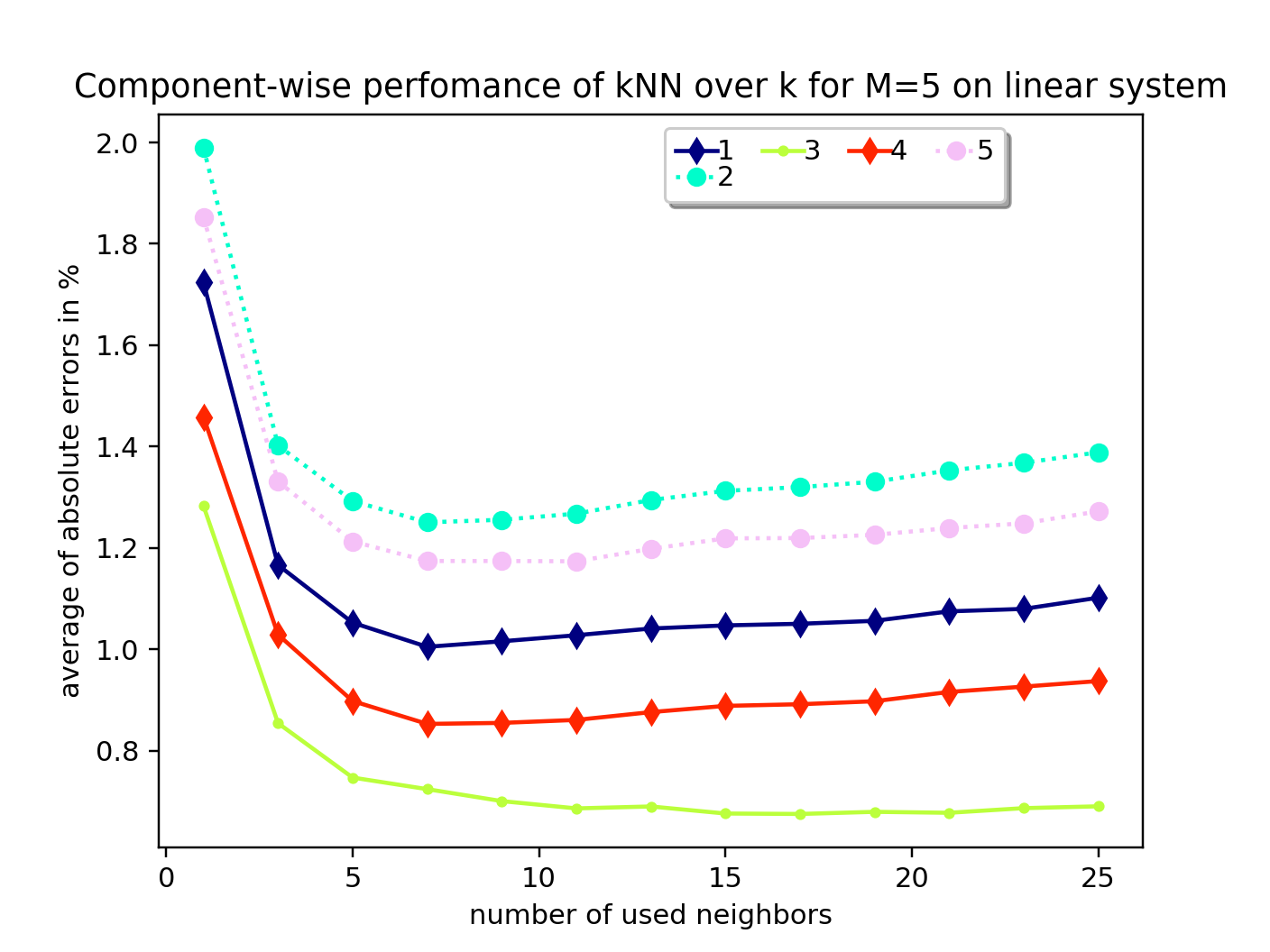}
 \caption{\label{fig:kNN_performance_02} The component-wise performance $\ell_1$ error in percent (Eq.~\eqref{eq:l1error}) of kNN in a linear low dimensional system with $L=7, M=5$. This shows individual views with averaged absolute deviations unlike Fig.~\ref{fig:kNN_performance_01}. The difference between individual performances are stochastic in nature depending on generated samples with the realized random matrix \mm{H}. The legends indicate component indices.}
\end{figure}

We first performed the linear low dimensional cases where $L=7, M=5, \sigma=0.005$ and \mm{H} is generated according to the standard Gaussian distribution for its entries. The true composition data is generated according to the uniform distribution and it is used to synthesize the observation following Eq.~\ref{eq:linearModel_noiseless} with additive noise. As discussed, the optimal performance is observed in the mid-range of $k=9$ or 11, shown in Fig.~\ref{fig:kNN_performance_01} and \ref{fig:kNN_performance_02} evaluated on the test data. 

\subsubsection{kNN in a linear high dimension}
\label{app:kNN_lin_highdim}

\begin{figure} [ht]
 \centering %  trim={<left> <lower> <right> <upper>}
 \includegraphics[width=4in,trim={0cm 0cm 0 0cm},clip]{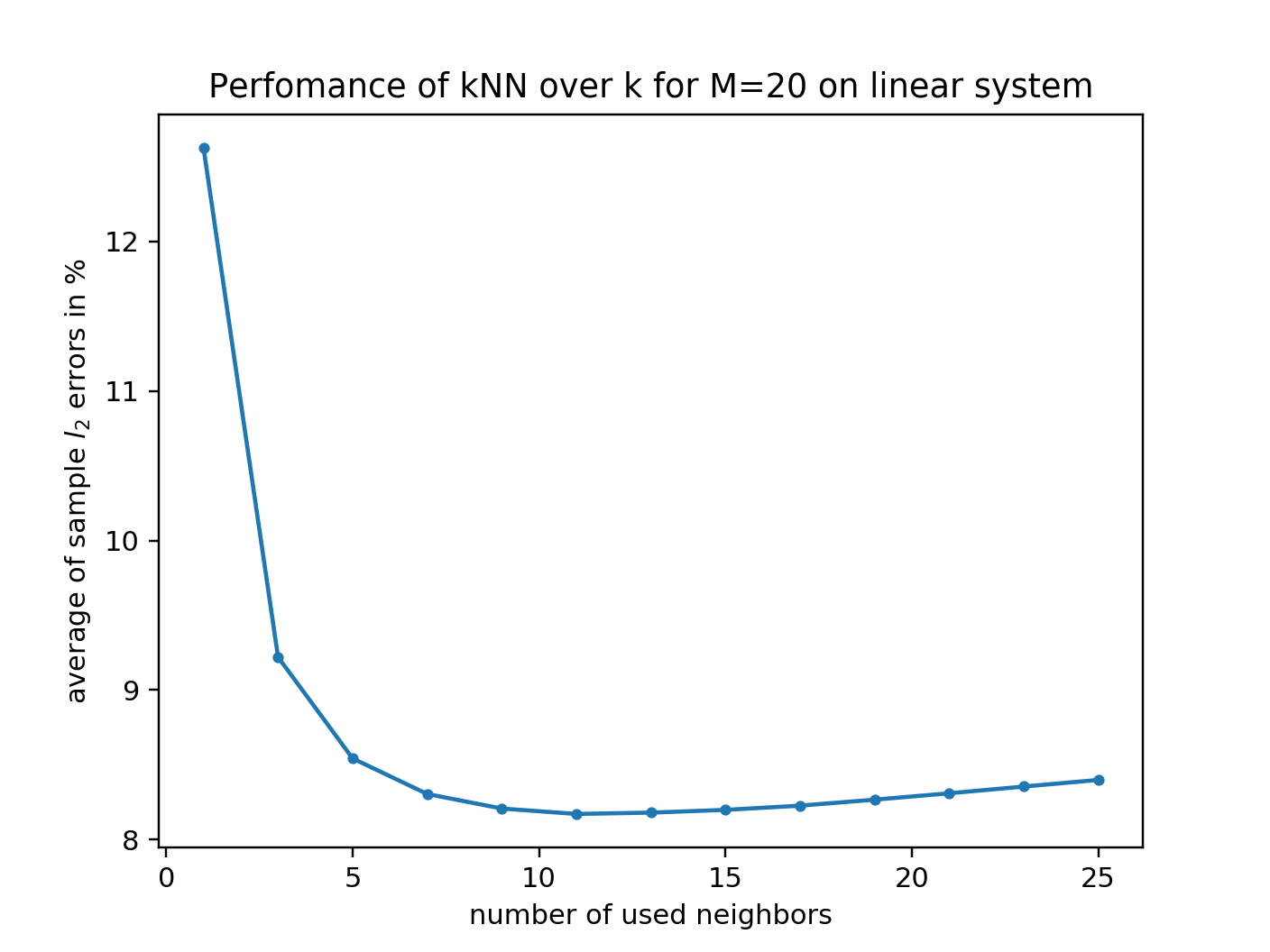}
 \caption{\label{fig:kNN_performance_M20_01} The average of $\ell_2$ errors in percent (Eq.~\eqref{eq:l2error}) as the performance of kNN in a linear high dimensional system with $M=20, L=1000$. Compared to the overall error in the low dimensional case in Fig.~\ref{fig:kNN_performance_01}, this shows an increased error level.}
 \includegraphics[width=4in,trim={0cm 0cm 0 0cm},clip]{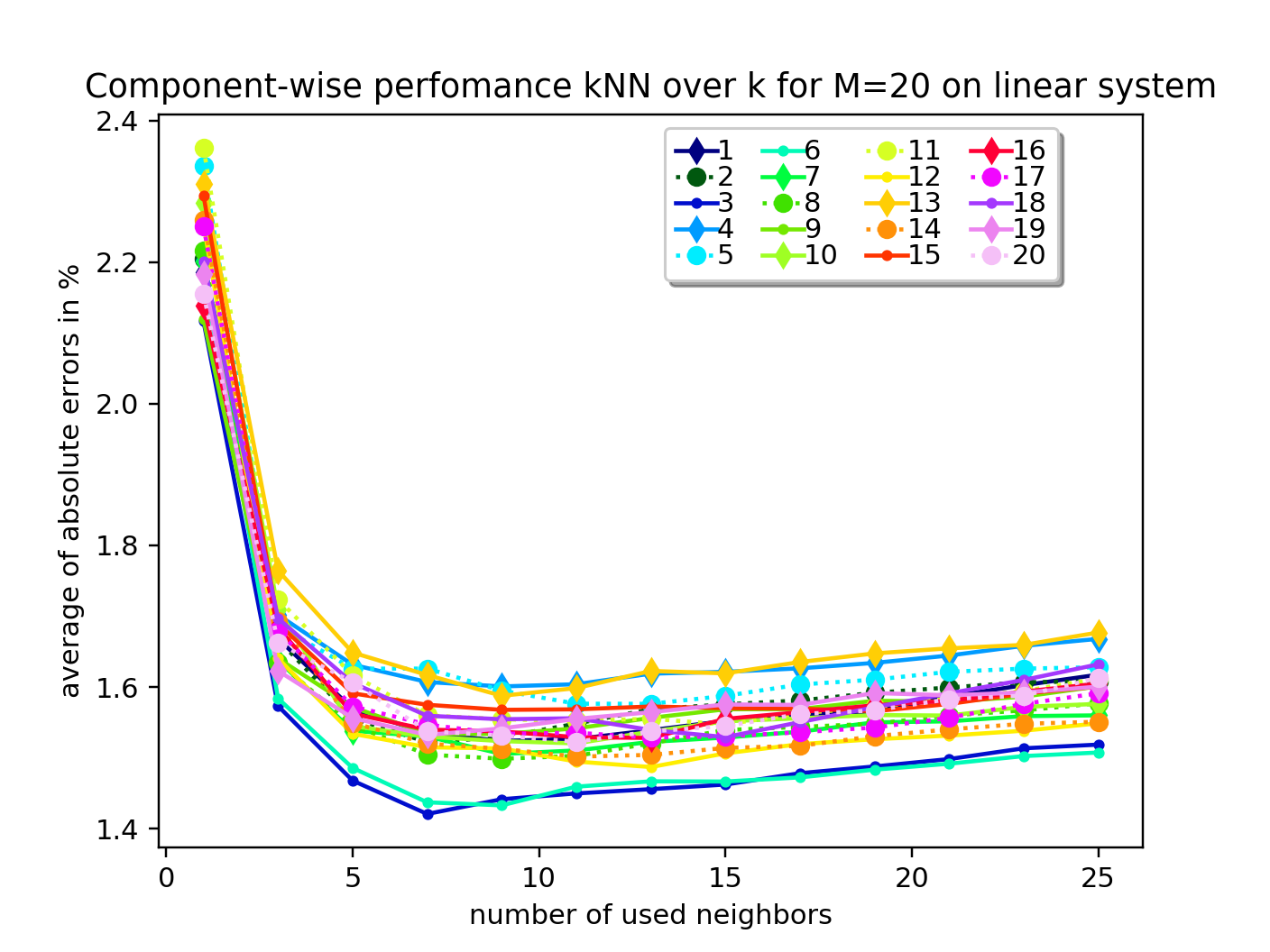}
 \caption{\label{fig:kNN_performance_M20_02} The component-wise performance $\ell_1$ error in percent (Eq.~\eqref{eq:l1error}) of kNN in a linear high dimensional system with $M=20, L=1000$. This shows individual views with averaged absolute deviations (AAD) unlike Fig.~\ref{fig:kNN_performance_M20_01}. Compared to the individual AADs in the low dimensional case in Fig.~\ref{fig:kNN_performance_02}, the ADDs here are comparable but the spread of individual performances is reduced with the much larger observation dimension $L$ and $M$. The comparable individual AADs leads to the larger overall $\ell_2$ error level in Figures \ref{fig:kNN_performance_M20_01} with $M=20$ comparing  Fig.~\ref{fig:kNN_performance_01}   with $M=5$.  The legends indicate component indices.}
\end{figure}

The high dimensional linear case was experimented with the same setting as the above except for $M=20, L=1000$. 
As seen in the low dimensional case, the optimal performance is observed in the mid-range of $k=9$ or 11, shown in Fig.~\ref{fig:kNN_performance_M20_01} and \ref{fig:kNN_performance_M20_02}. The individual error is comparable but the overall error increased compared to the previous linear low dimension case in Appendix.~\ref{app:kNN_lin_lowdim}.

\subsubsection{kNN in a nonlinear high dimension with obfuscating variables}
\label{app:kNN_nonlin_highdim}

\begin{figure} [ht]
 \centering %  trim={<left> <lower> <right> <upper>}
 \includegraphics[width=4in,trim={0cm 0cm 0 0cm},clip]{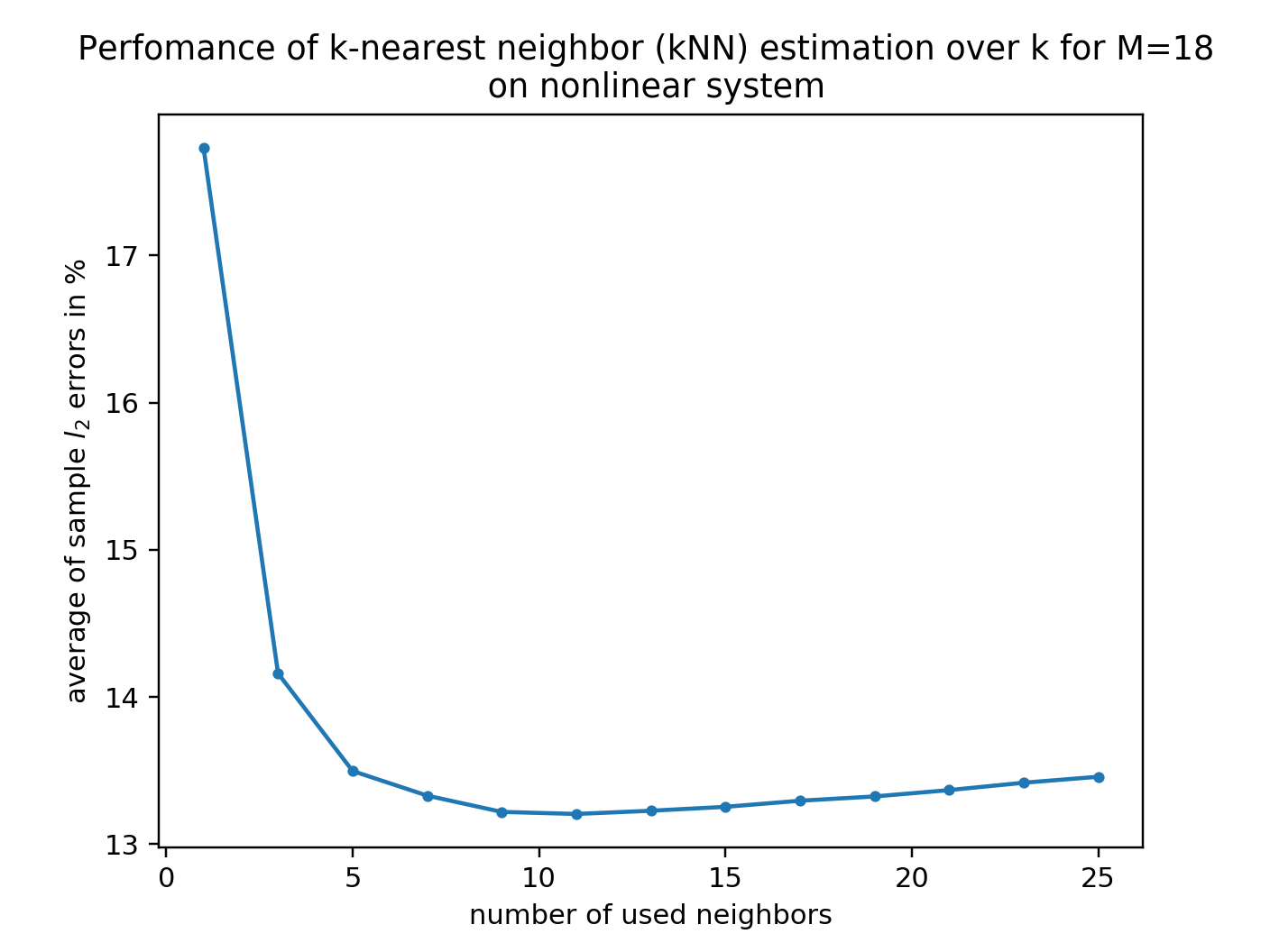}
 \caption{\label{fig:kNN_performance_nonlin_01} The average of $\ell_2$ errors in percent (Eq.~\eqref{eq:l2error}) as the performance of kNN in a nonlinear high dimensional system with $L=1000$, 2 obfuscating variables and effective $M=18$ (the number of components of interest or modelled). The error is larger due to the nonlinearity of the given system than that in Fig.~\ref{fig:kNN_performance_M20_01}. }
 \includegraphics[width=4in,trim={0cm 0cm 0 0cm},clip]{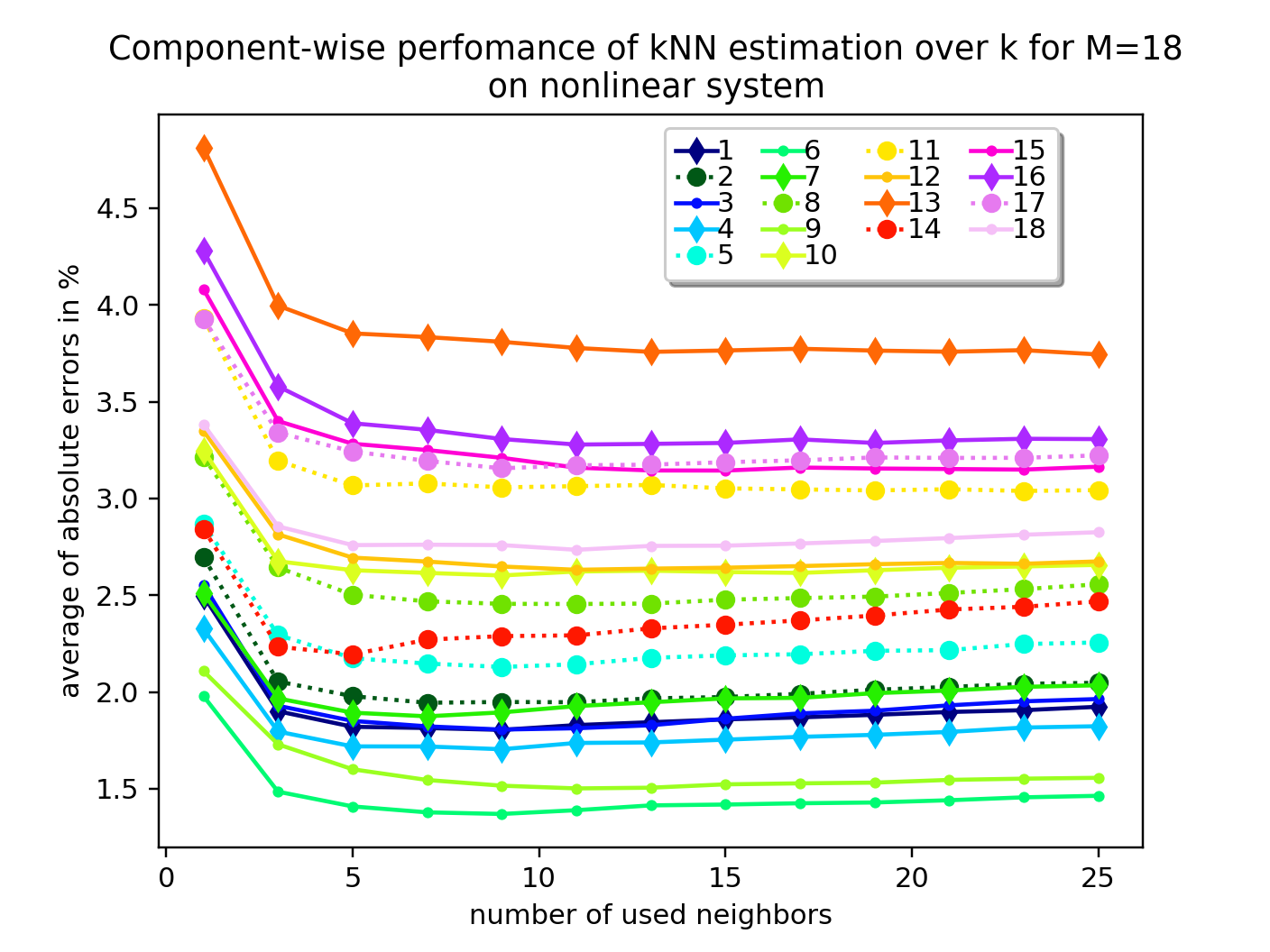}
 \caption{\label{fig:kNN_performance_nonlin_02} The component-wise performance $\ell_1$ error in percent (Eq.~\eqref{eq:l1error}) of kNN in a linear high dimensional system. This shows individual views with averaged absolute deviations (AAD) unlike Fig.~\ref{fig:kNN_performance_nonlin_01}. Compared to the individual AADs in Fig.~\ref{fig:kNN_performance_02}, the ADDs here are comparable but the spread of individual performances is reduced with the much larger observation dimension $L$ and $M$.  The individual errors are larger due to the nonlinearity of the given system than that in Fig.~\ref{fig:kNN_performance_M20_02}.  The legends indicate component indices.}
\end{figure}

We then simulated the high dimensional nonlinear case with the same setting as the above high dimensional linear  case except for $M=20, L=1000$, the nonlinear system defined in Appendix.~\ref{app:highdim_nonlin}, 2 obfuscating variables out of $M=20$ components, and using uniformly distributed compositional vectors. 
As seen in previous cases, the optimal performance is observed in the mid-range of $k=9$ or 11, shown in Fig.~\ref{fig:kNN_performance_nonlin_01} and \ref{fig:kNN_performance_nonlin_02}. The larger errors than those in the previous high dimensional linear case are evident due to the nonlinearity of the defined system, based on the evaluation on the test data.

\subsubsection{kNN in a nonlinear high dimension with obfuscating variables and mixture models}
\label{app:kNN_nonlin_highdim_mix}

\begin{figure} [ht]
 \centering %  trim={<left> <lower> <right> <upper>}
 \includegraphics[width=4in,trim={0cm 0cm 0 0cm},clip]{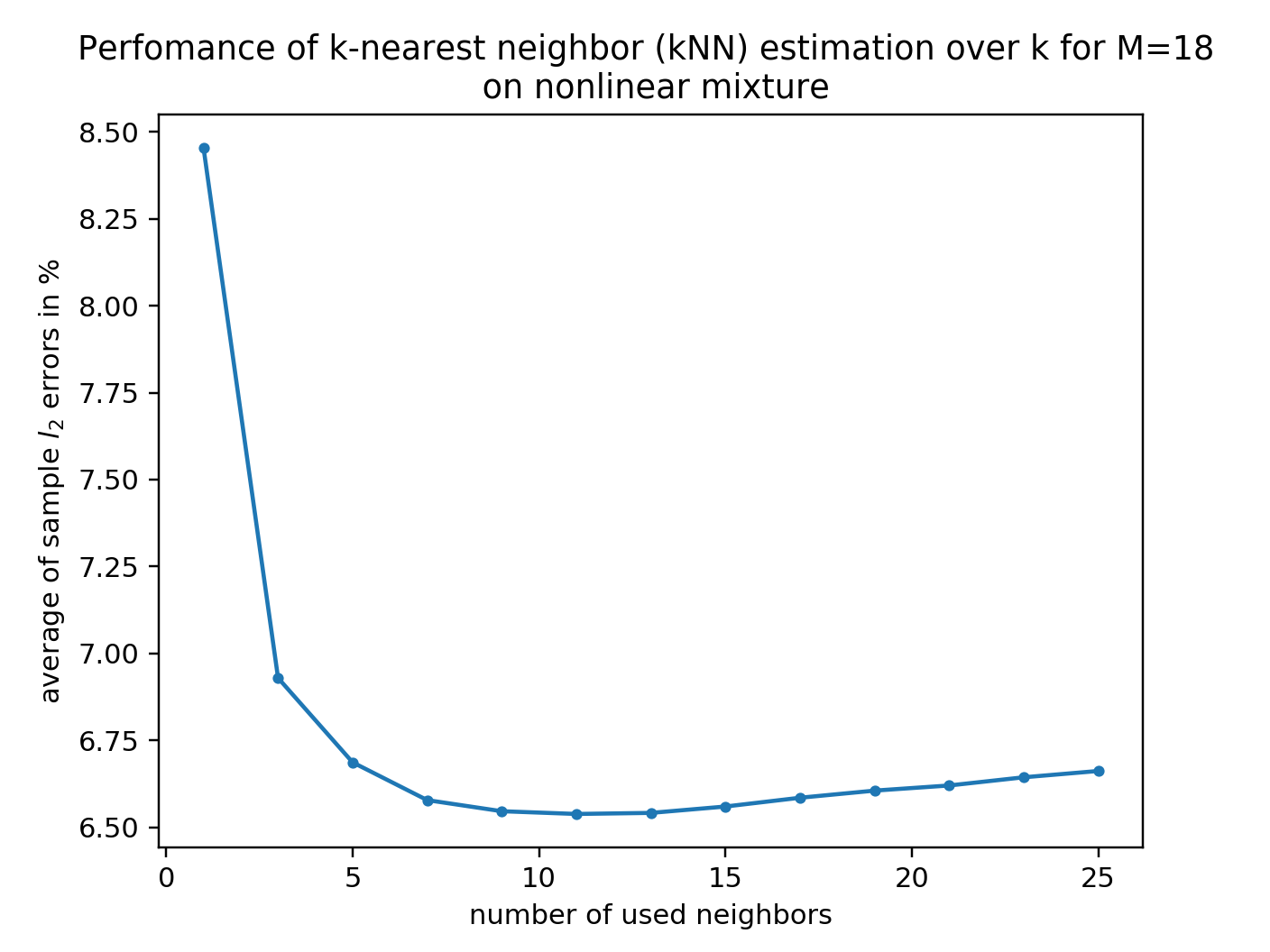}
 \caption{\label{fig:kNN_performance_nonlin_mix_01} The average of $\ell_2$ errors in percent (Eq.~\eqref{eq:l2error}) as the performance of kNN in a nonlinear high dimensional system with $L=1000$, 2 obfuscating variables so effective $M=18$ (the number of components of interest or modelled), and the mixture model. The increased error due to the nonlinearity of the given system is reduced because of the concentration or dense neighborhood compared to the previous case in Fig.~\ref{fig:kNN_performance_nonlin_01}. }
 \includegraphics[width=4in,trim={0cm 0cm 0 0cm},clip]{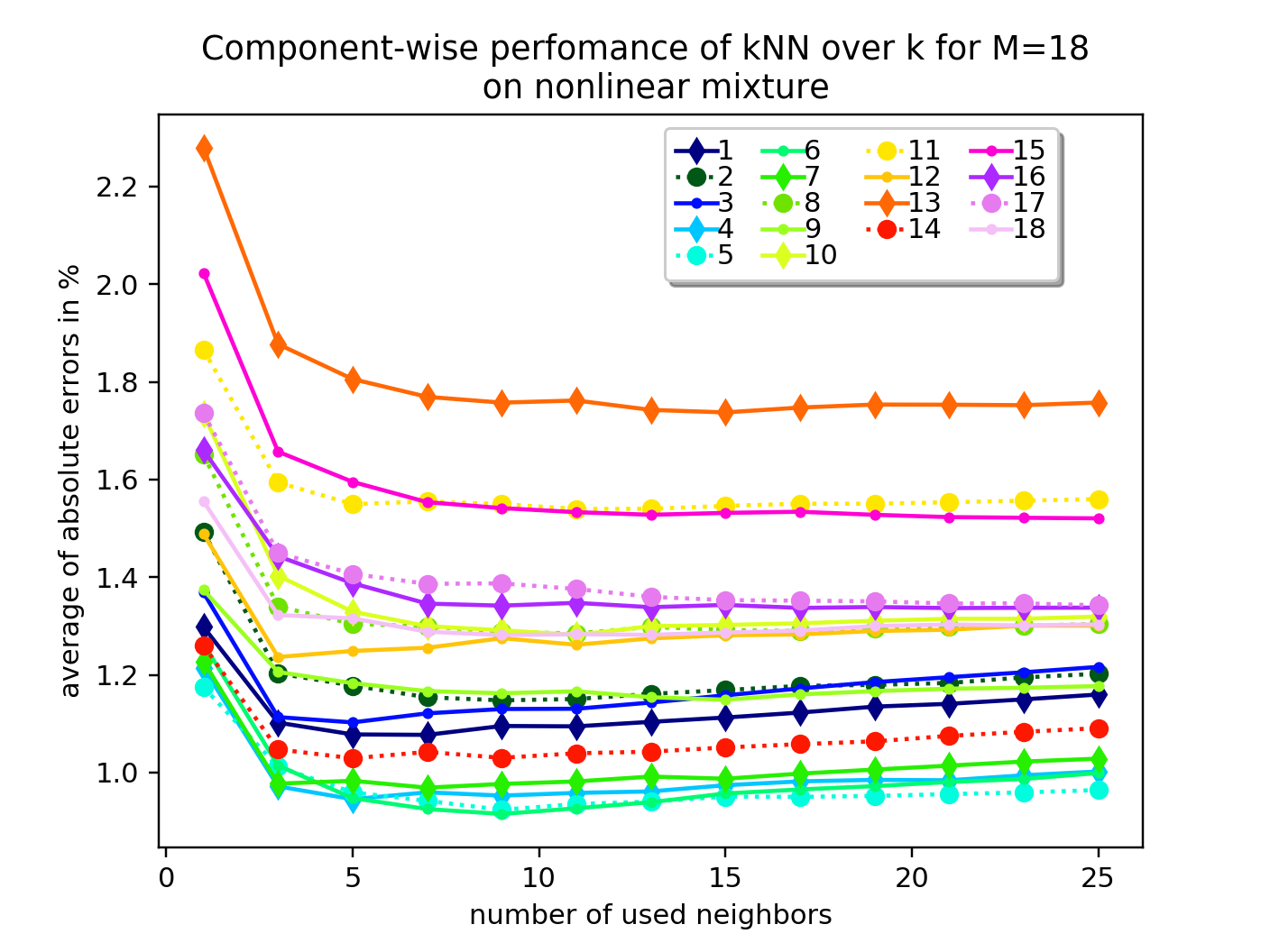}
 \caption{\label{fig:kNN_performance_nonlin_mix_02} The component-wise performance $\ell_1$ error in percent (Eq.~\eqref{eq:l1error}) of kNN in a nonlinear high dimensional system with $L=1000$, 2 obfuscating variables so effective $M=18$, and the mixture model.  This shows individual views with averaged absolute deviations (AAD) unlike Fig.~\ref{fig:kNN_performance_nonlin_mix_01}. Compared to the individual AADs in Fig.~\ref{fig:kNN_performance_nonlin_02}, the relative trends of ADDs are similar but the error levels are reduced due to the neighborhood concentration.  The legends indicate component indices.  }
\end{figure}

This case adds more complexity to the high dimensional nonlinear case with the mixtures defined in Appendix.~\ref{app:highdim_nonlin}, so the compositional vectors are sampled from the mixture of three truncated approximate Gaussian distributions and uniform distribution.
As seen in the previous cases, the optimal performance is observed in the mid-range of $k=9$ or 11, shown in Fig.~\ref{fig:kNN_performance_nonlin_mix_01} and \ref{fig:kNN_performance_nonlin_mix_02}. Because the three Gaussian distributions are much more concentrated, thus having dense neighborhood, than the uniform distribution, which is the most scattered distribution as the worst case, the performance is improved significantly compared to the previous case of having only uniformly distributed samples. 

\clearpage
\newpage

\bibliographystyle{IEEEtran}
\bibliography{bibl05}

\section*{Author Bio}
{\bf Se Un Park} is Senior Data Scientist at Schlumberger. His research interests include stochastic signal processing, variational Bayesian methods, blind deconvolution, machine learning, and prognostics and health management, and mineralogy characterization for oil field reservoirs. He holds a PhD degree in Electrical Engineering and Computer Science from the University of Michigan, Ann Arbor, MI, USA.

\end{document}